\documentclass[10pt,twocolumn,letterpaper]{article}

\usepackage[pagenumbers]{cvpr}

\usepackage{microtype}

\usepackage{amsmath,amssymb}
\DeclareMathSizes{7}{7}{6}{5}
\DeclareMathSizes{8}{8}{7}{5}
\usepackage{booktabs}
\usepackage{multirow}
\usepackage{xspace}
\usepackage{graphicx}
\usepackage{xcolor}
\usepackage{tabularx}

\newcommand{\method}{ShelfKV\xspace}
\newcommand{\desk}{desk\xspace}
\newcommand{\shelf}{shelf\xspace}

\newcommand{\A}{\mathcal{A}}   %
\newcommand{\V}{\mathcal{V}}   %
\newcommand{\R}{\mathcal{R}}   %

\usepackage{tikz}
\usetikzlibrary{arrows.meta,positioning,fit,backgrounds,calc,decorations.pathreplacing,patterns}
\usepackage{pgfplots}
\usepgfplotslibrary{groupplots}
\pgfplotsset{compat=1.18}
\pgfplotsset{/pgf/number format/assume math mode=true}%
\newlength{\figpx}%
\definecolor{fOut}{HTML}{3A3F45}      %
\definecolor{fBub}{HTML}{E9EEF5}      %
\definecolor{fBubD}{HTML}{5F6670}
\definecolor{fTile}{HTML}{6E747B}     %
\definecolor{fDash}{HTML}{9AA0A6}     %
\definecolor{fGreen}{HTML}{2E7D32}
\definecolor{fPurple}{HTML}{6A3D9A}
\definecolor{fSkin}{HTML}{F6D8BF}
\definecolor{fHair}{HTML}{5B3A2A}
\definecolor{fShirt}{HTML}{8FCBF0}
\definecolor{fMouth}{HTML}{E05A5A}
\definecolor{fBlush}{HTML}{F4A9A0}
\definecolor{fRobot}{HTML}{D9DCE0}
\definecolor{fScreen}{HTML}{3A3B3F}
\definecolor{fWood}{HTML}{F3C98B}
\definecolor{fWoodD}{HTML}{6B4423}
\definecolor{fDesk}{HTML}{F8E2BD}
\definecolor{fLabel}{HTML}{E8E8E8}
\definecolor{fPaper}{HTML}{F1F2F3}
\definecolor{fArrow}{HTML}{4A4D52}
\definecolor{fRed}{HTML}{E53935}
\definecolor{fCheck}{HTML}{3E9B2F}
\definecolor{fSpark}{HTML}{F5B21B}
\tikzset{
  ol/.style={draw=fOut, line width=2.6\figpx, line join=round, line cap=round},
  thin ol/.style={draw=fOut, line width=1.8\figpx, line join=round, line cap=round},
  tileb/.style={draw=fTile, line width=2.4\figpx, fill=white, rounded corners=7\figpx},
  deleted/.style={draw=fDash, line width=2.4\figpx, dash pattern=on 9\figpx off 6\figpx, rounded corners=9\figpx},
}
\newcommand{\fBubble}[6]{%
  \filldraw[fill=fBub, draw=fBubD, line width=2.6\figpx, rounded corners=16\figpx] (#1,#2) rectangle (#3,#4);
  \pgfmathsetmacro{\ym}{(#2+#4)/2}%
  \fill[fBub] (#1+1.4,\ym-11) -- (#5,#6) -- (#1+1.4,\ym+9) -- cycle;
  \draw[draw=fBubD, line width=2.6\figpx, line join=round] (#1,\ym-11) -- (#5,#6) -- (#1,\ym+9);}
\newcommand{\fGirl}[2]{\begin{scope}[shift={(#1,#2)}]
  \filldraw[fill=fHair, ol] (-38,8) .. controls (-46,-46) and (46,-46) .. (38,8) -- (41,40) .. controls (30,47) and (24,44) .. (20,36) -- (-20,36) .. controls (-24,44) and (-30,47) .. (-41,40) -- cycle;
  \filldraw[fill=fShirt, ol] (-34,62) .. controls (-32,38) and (32,38) .. (34,62) -- cycle;
  \filldraw[fill=fSkin, ol] (-7,30) rectangle (7,44);
  \filldraw[fill=fSkin, ol] (0,6) ellipse (26 and 28);
  \fill[fHair] (-27,4) .. controls (-30,-28) and (18,-36) .. (27,-2) .. controls (12,-12) and (-2,-16) .. (-7,-10) .. controls (-12,-3) and (-20,0) .. (-27,4) -- cycle;
  \fill[fBlush] (-16,15) circle (4.5); \fill[fBlush] (16,15) circle (4.5);
  \fill[fOut] (-9,6) ellipse (2.6 and 3.8); \fill[fOut] (9,6) ellipse (2.6 and 3.8);
  \filldraw[fill=fMouth, thin ol] (-7,15) .. controls (-6,27) and (6,27) .. (7,15) -- cycle;
  \draw[draw=fSpark!70!orange, line width=3.2\figpx, line cap=round] (30,-34) -- (34,-46) (36,-26) -- (47,-32) (37,-15) -- (49,-15);
\end{scope}}
\newcommand{\fRobot}[3]{\begin{scope}[shift={(#1,#2)}]
  \filldraw[fill=fRobot, ol] (-34,74) .. controls (-34,40) and (-20,30) .. (0,30) .. controls (20,30) and (34,40) .. (34,74) -- cycle;
  \filldraw[fill=fRobot, ol] (-9,26) rectangle (9,38);
  \draw[ol] (0,-36) -- (0,-50); \filldraw[fill=fRobot, ol] (0,-57) circle (7.5);
  \filldraw[fill=fRobot, ol] (-41,-3) circle (10); \filldraw[fill=fRobot, ol] (41,-3) circle (10);
  \filldraw[fill=fRobot, ol, rounded corners=17\figpx] (-37,-37) rectangle (37,32);
  \filldraw[fill=fScreen, draw=fScreen, rounded corners=12\figpx] (-27,-24) rectangle (27,19);
  \ifcase#3
    \fill[white] (-11,-5) circle (5.5); \fill[white] (11,-5) circle (5.5);
    \draw[white, line width=2.6\figpx, line cap=round] (-6,8) .. controls (-2,11) and (2,11) .. (6,8);
  \or
    \draw[white, line width=3.2\figpx, line cap=round] (-17,-2) .. controls (-14,-10) and (-8,-10) .. (-5,-2);
    \draw[white, line width=3.2\figpx, line cap=round] (5,-2) .. controls (8,-10) and (14,-10) .. (17,-2);
    \draw[white, line width=3.2\figpx, line cap=round] (-7,6) .. controls (-3,12) and (3,12) .. (7,6);
  \or
    \fill[white] (-11,-3) ellipse (4.5 and 5.5); \fill[white] (11,-3) ellipse (4.5 and 5.5);
    \draw[white, line width=2.6\figpx, line cap=round] (-17,-11) -- (-6,-16) (17,-11) -- (6,-16);
    \draw[white, line width=2.6\figpx, line cap=round] (-6,12) .. controls (-2,8) and (2,8) .. (6,12);
  \fi
\end{scope}}
\newcommand{\fGlyph}[5]{%
  \pgfmathsetmacro{\gx}{#1}\pgfmathsetmacro{\gy}{#2}\pgfmathsetmacro{\gxe}{#3}\pgfmathsetmacro{\gye}{#4}\pgfmathsetmacro{\gw}{\gxe-\gx}\pgfmathsetmacro{\gh}{\gye-\gy}%
  \ifcase#5\or
    \draw[fOut, line width=3\figpx, line cap=round] (\gx+0.62*\gw,\gy+0.14*\gh) -- (\gx+0.62*\gw,\gye+3);
    \draw[fOut, line width=3\figpx, line cap=round] (\gx+0.50*\gw,\gy+0.36*\gh) -- (\gx+0.62*\gw,\gy+0.36*\gh) (\gx+0.50*\gw,\gy+0.80*\gh) -- (\gx+0.62*\gw,\gy+0.80*\gh);
  \or
    \draw[fGreen, line width=3.4\figpx, line join=round] (\gx+0.14*\gw,\gy+0.80*\gh) -- (\gx+0.50*\gw,\gy+0.56*\gh) -- (\gx+0.86*\gw,\gy+0.30*\gh);
    \foreach \u/\v in {0.14/0.80, 0.50/0.56, 0.86/0.30} \filldraw[fill=white, draw=fGreen, line width=2.6\figpx] (\gx+\u*\gw,\gy+\v*\gh) circle (5.5);
  \or
    \draw[fPurple, line width=3.4\figpx, line join=round] (\gx+0.14*\gw,\gy+0.26*\gh) -- (\gx+0.50*\gw,\gy+0.50*\gh) -- (\gx+0.86*\gw,\gy+0.76*\gh);
    \foreach \u/\v in {0.14/0.26, 0.50/0.50, 0.86/0.76} \filldraw[fill=white, draw=fPurple, line width=2.6\figpx] (\gx+\u*\gw,\gy+\v*\gh) circle (5.5);
  \or
    \draw[fOut, line width=3\figpx, line cap=round, line join=round] (\gx+0.62*\gw,\gy-3) -- (\gx+0.62*\gw,\gy+0.72*\gh) -- (\gxe+3,\gy+0.72*\gh);
  \or
    \draw[fOut, line width=3\figpx, line cap=round] (\gx-3,\gy+0.72*\gh) -- (\gxe+3,\gy+0.72*\gh);
    \draw[fOut, line width=3\figpx, line cap=round] (\gx+0.30*\gw,\gy+0.72*\gh) -- (\gx+0.30*\gw,\gy+0.58*\gh) (\gx+0.80*\gw,\gy+0.72*\gh) -- (\gx+0.80*\gw,\gy+0.58*\gh);
  \fi}
\newcommand{\fTile}[6][fTile]{%
  \filldraw[fill=white, draw=#1, line width=2.6\figpx, rounded corners=7\figpx] (#2,#3) rectangle (#4,#5);
  \begin{scope}\clip[rounded corners=7\figpx] (#2,#3) rectangle (#4,#5); \fGlyph{#2}{#3}{#4}{#5}{#6}\end{scope}}
\newcommand{\fTrash}[4]{%
  \draw[deleted] (#1,#2) rectangle (#3,#4);
  \pgfmathsetmacro{\cx}{(#1+#3)/2}\pgfmathsetmacro{\cy}{(#2+#4)/2}%
  \begin{scope}[shift={(\cx,\cy)}]
    \draw[fDash!90!black, line width=2.4\figpx, line join=round, line cap=round, fill=white] (-13,-12) -- (-10,19) -- (10,19) -- (13,-12);
    \draw[fDash!90!black, line width=2.4\figpx, line cap=round] (-16,-13) -- (16,-13) (-5,-13) -- (-5,-18) -- (5,-18) -- (5,-13);
    \draw[fDash!90!black, line width=2.2\figpx, line cap=round] (-4.5,-1) -- (4.5,9) (4.5,-1) -- (-4.5,9);
  \end{scope}}
\newcommand{\fCard}[4]{%
  \filldraw[fill=fPaper, draw=fTile, line width=2.4\figpx, rounded corners=5\figpx] (#1,#2) rectangle (#3,#4);
  \pgfmathsetmacro{\cw}{(#3)-(#1)}\pgfmathsetmacro{\ch}{(#4)-(#2)}%
  \foreach \v in {0.30,0.50,0.70} \draw[fDash, line width=2.4\figpx, line cap=round] (#1+0.22*\cw,#2+\v*\ch) -- (#3-0.22*\cw,#2+\v*\ch);}
\newcommand{\fBrace}[4]{\draw[#4, line width=2.4\figpx, line cap=round, decorate, decoration={brace, amplitude=9\figpx, raise=0pt}] (#1,#3) -- (#2,#3);}

\definecolor{bpOurs}{HTML}{0072B2}   %
\definecolor{bpBase}{HTML}{D55E00}   %
\definecolor{bpVis}{HTML}{009E73}    %
\definecolor{bpText}{HTML}{7F7F7F}   %

\usepackage{xcolor}
\usepackage{colortbl}

\definecolor{sktHeader}{HTML}{E8EDF2}
\definecolor{sktGroup}{HTML}{F3F5F7}
\definecolor{sktHighlight}{HTML}{E4F0F8}
\definecolor{sktAccent}{HTML}{0072B2}
\usepackage{array}
\definecolor{cvprblue}{rgb}{0.21,0.49,0.74}
\usepackage[pagebackref,breaklinks,colorlinks,allcolors=cvprblue]{hyperref}
\usepackage[capitalize]{cleveref}

\def\confName{CVPR}
\def\confYear{2027}

\title{Keep the Reasoning, Read the Image:\\On-Demand Visual KV for Multi-Turn Reasoning VLMs}

\author{
Ziyi Wang$^{1,*}$\quad
Xiyue Zhang$^{1,*}$\quad
Bing Hu$^{1}$\quad
Guangmin Song$^{2}$\quad
Chenyi Zhang$^{3}$\quad
Li Wang$^{4}$\\[0.2em]
{\normalsize
$^{1}$Zhejiang University\quad
$^{2}$Zhejiang Sci-Tech University\quad
$^{3}$University of Canterbury\quad
$^{4}$Wenzhou Business College}
}

\makeatletter
\def\@maketitle{%
  \newpage
  \null
  \vskip 0pt
  \begin{center}%
    {\Large \bf \@title \par}%
    \vspace*{10pt}%
    {\large
      \lineskip .5em
      \begin{tabular}[t]{c}\@author\end{tabular}\par}%
    \vskip .5em
    \vspace*{12pt}%
  \end{center}}
\makeatother

\begin{document}
\twocolumn[{%
\renewcommand\twocolumn[1][]{#1}%
\maketitle
\vspace{\dimexpr-1.2em+7pt\relax}
\begin{center}
  \captionsetup{type=figure}
  \centering
  \setlength{\figpx}{0.000362\textwidth}%
  \begin{minipage}[b]{0.695\textwidth}
    \centering
    \begin{tikzpicture}[x=\figpx, y=-\figpx, every node/.style={inner sep=0pt, align=center}]
      \draw[fDash!60, line width=2\figpx] (520,12) -- (520,808);
\draw[fDash!60, line width=2\figpx] (1196,12) -- (1196,808);
\fGirl{70}{128}
\fBubble{153}{74}{494}{182}{122}{130}
\filldraw[fill=white, draw=fTile, line width=2.6\figpx, rounded corners=16\figpx] (35,200) rectangle (495,355);
\begin{scope}
  \draw[fOut, line width=3\figpx, line cap=round, line join=round] (107,222) -- (107,338) -- (258,338) (330,222) -- (330,338) -- (480,338);
  \foreach \x in {107,330} \draw[fOut, line width=3\figpx, line cap=round] (\x-7,240) -- (\x,240) (\x-7,305) -- (\x,305);
  \foreach \x in {150,200,250} \draw[fOut, line width=2.6\figpx] (\x-40,338) -- (\x-40,331);
  \foreach \x in {372,422,472} \draw[fOut, line width=2.6\figpx] (\x-40,338) -- (\x-40,331);
  \draw[fGreen, line width=3.4\figpx, line join=round] (117,307) -- (148,289) -- (181,276) -- (214,265) -- (247,252);
  \foreach \p in {(117,307),(148,289),(181,276),(214,265),(247,252)} \filldraw[fill=white, draw=fGreen, line width=2.6\figpx] \p circle (5.5);
  \draw[fPurple, line width=3.4\figpx, line join=round] (352,248) -- (389,268) -- (426,286) -- (462,303);
  \foreach \p in {(352,248),(389,268),(426,286),(462,303)} \filldraw[fill=white, draw=fPurple, line width=2.6\figpx] \p circle (5.5);
\end{scope}
\filldraw[fill=fPaper, ol] (150,398) rectangle (482,505);
\filldraw[fill=fPaper, ol] (150,451.5) ellipse (17 and 53.5);
\foreach \y/\a/\b in {418/190/300, 418/320/450, 486/190/260, 486/280/430} \draw[fDash, line width=3\figpx, line cap=round] (\a,\y) -- (\b,\y);
\fRobot{86}{432}{1}
\draw[fOut, line width=11\figpx, line cap=round] (108,482) .. controls (124,476) and (136,468) .. (146,458);
\draw[fRobot, line width=6\figpx, line cap=round] (108,482) .. controls (124,476) and (136,468) .. (146,458);
\draw[fSpark, line width=3.4\figpx, line cap=round] (30,375) -- (40,388) (20,393) -- (34,398);
\foreach \c/\k in {0/1,1/2,2/1,3/3} {\pgfmathsetmacro{\xx}{63+\c*101}\fTile{\xx}{528}{\xx+93}{593}{\k}}
\foreach \c/\k in {0/1,1/2,2/1,3/3} {\pgfmathsetmacro{\xx}{63+\c*101}\fTile{\xx}{600}{\xx+93}{665}{\k}}
\foreach \c/\k in {0/4,1/5,2/4,3/5} {\pgfmathsetmacro{\xx}{63+\c*101}\fTile{\xx}{672}{\xx+93}{737}{\k}}
\filldraw[fill=fPaper, draw=fTile, line width=2.6\figpx, rounded corners=12\figpx] (40,755) rectangle (484,802);
\foreach \x in {62,138,214,290,366} \draw[fDash, line width=3.4\figpx, line cap=round] (\x,772) -- (\x+56,772);
\foreach \x in {62,138,214,290} \draw[fDash, line width=3.4\figpx, line cap=round] (\x,787) -- (\x+56,787);
\foreach \x in {430,443,456} \fill[fDash] (\x,787) circle (2.6);
\fGirl{590}{128}
\fBubble{673}{74}{1062}{182}{642}{130}
\fBrace{565}{860}{306}{fDash!90!black}
\fBrace{877}{963}{306}{bpBase}
\foreach \r in {0,1} \foreach \c in {0,1,2} {\pgfmathsetmacro{\xx}{565+\c*104}\pgfmathsetmacro{\yy}{318+\r*104}\fTrash{\xx}{\yy}{\xx+88}{\yy+88}}
\foreach \c in {0,1} {\pgfmathsetmacro{\xx}{565+\c*104}\fTrash{\xx}{526}{\xx+88}{614}}
\fTrash{773}{526}{861}{614}
\fTile[bpBase]{877}{318}{963}{404}{3}
\fTile[bpBase]{877}{422}{963}{508}{3}
\draw[fOut, line width=2.6\figpx, dash pattern=on 9\figpx off 6\figpx, rounded corners=9\figpx] (877,526) rectangle (965,614);
\begin{scope}[shift={(921,570)}]
  \draw[fDash!90!black, line width=2.4\figpx, line join=round, line cap=round, fill=white] (-13,-12) -- (-10,19) -- (10,19) -- (13,-12);
  \draw[fDash!90!black, line width=2.4\figpx, line cap=round] (-16,-13) -- (16,-13) (-5,-13) -- (-5,-18) -- (5,-18) -- (5,-13);
  \draw[fDash!90!black, line width=2.2\figpx, line cap=round] (-4.5,-1) -- (4.5,9) (4.5,-1) -- (-4.5,9);
\end{scope}
\draw[fOut, line width=2.2\figpx, dash pattern=on 4\figpx off 5\figpx] (965,532) -- (1012,470) (965,608) -- (1028,606);
\filldraw[fill=white, draw=fOut, line width=2.6\figpx, dash pattern=on 4\figpx off 5\figpx] (1090,535) circle (95);
\draw[fOut, line width=3.4\figpx, line cap=round, line join=round] (1102,458) -- (1102,602) -- (1150,602);
\draw[fOut, line width=3.4\figpx, line cap=round] (1094,480) -- (1110,480) (1094,560) -- (1110,560);
\fRobot{683}{728}{2}
\node[font=\bfseries, text=fOut, scale=1.25] at (748,672) {?};
\fBubble{805}{693}{928}{766}{775}{735}
\draw[fRed, line width=8\figpx, line cap=round] (958,710) -- (994,746) (994,710) -- (958,746);
\fGirl{1266}{128}
\fBubble{1349}{74}{1738}{182}{1318}{130}
\fill[fLabel, rounded corners=14\figpx] (1205,226) rectangle (1495,334);
\fill[fLabel, rounded corners=14\figpx] (1628,226) rectangle (1908,334);
\filldraw[fill=fWood, draw=fWoodD, line width=2.6\figpx, rounded corners=6\figpx] (1210,346) rectangle (1230,612);
\filldraw[fill=fWood, draw=fWoodD, line width=2.6\figpx, rounded corners=6\figpx] (1470,346) rectangle (1490,612);
\filldraw[fill=fWood!45, draw=fWoodD, line width=2.2\figpx] (1230,359) rectangle (1470,598);
\foreach \y in {352,428,508,590} \filldraw[fill=fWood, draw=fWoodD, line width=2.6\figpx, rounded corners=3\figpx] (1216,\y-7) rectangle (1484,\y+7);
\foreach \c/\k in {0/1,1/2,2/1,3/3} {\pgfmathsetmacro{\xx}{1238+\c*58}\fTile{\xx}{364}{\xx+51}{419}{\k}}
\foreach \c/\k in {0/1,1/2,2/1,3/3} {\pgfmathsetmacro{\xx}{1238+\c*58}\fTile{\xx}{438}{\xx+51}{499}{\k}}
\foreach \c/\k in {0/4,1/5,2/4,3/5} {\pgfmathsetmacro{\xx}{1238+\c*58}\fTile{\xx}{518}{\xx+51}{581}{\k}}
\fill[fArrow, rounded corners=2\figpx] (1503,423) rectangle (1600,437);
\fill[fArrow, rounded corners=2\figpx] (1596,410) -- (1630,430) -- (1596,450) -- cycle;
\fTile[bpOurs]{1503}{462}{1555}{522}{1}
\fTile[bpOurs]{1563}{462}{1615}{522}{1}
\filldraw[fill=fWood, draw=fWoodD, line width=2.6\figpx] (1645,575) rectangle (1662,612) (1873,575) rectangle (1890,612);
\filldraw[fill=fDesk, draw=fWoodD, line width=2.6\figpx, rounded corners=14\figpx] (1628,378) rectangle (1908,566);
\filldraw[fill=fWood, draw=fWoodD, line width=2.6\figpx, rounded corners=5\figpx] (1622,556) rectangle (1914,580);
\fTile[bpOurs]{1738}{396}{1796}{462}{3}
\fCard{1804}{398}{1848}{456}
\fCard{1856}{398}{1900}{456}
\fCard{1792}{478}{1835}{530}
\fCard{1842}{478}{1885}{530}
\filldraw[fill=fWood!70, draw=fWoodD, line width=2.4\figpx, rounded corners=4\figpx] (1772,512) rectangle (1906,552);
\draw[bpOurs, line width=7\figpx, line cap=round] (1712,465) -- (1735,500);
\filldraw[fill=white, draw=bpOurs, line width=6\figpx] (1683,430) circle (40);
\draw[fOut, line width=3.2\figpx, line cap=round, line join=round] (1703,398) -- (1703,462);
\draw[fOut, line width=3.2\figpx, line cap=round] (1696,411) -- (1710,411) (1696,452) -- (1710,452);
\fRobot{1398}{728}{1}
\draw[fSpark, line width=3.4\figpx, line cap=round] (1450,664) -- (1456,678) (1466,670) -- (1474,682) (1474,690) -- (1488,692);
\fBubble{1517}{692}{1640}{765}{1487}{735}
\draw[fCheck, line width=9\figpx, line cap=round, line join=round] (1668,728) -- (1685,746) -- (1716,706);

      \node[anchor=west, font=\footnotesize\bfseries] at (26,34) {(a) Round 1: reason};
      \node[anchor=west, font=\footnotesize\bfseries] at (546,34) {Round 2: {\color{bpBase}retention policy}};
      \node[anchor=west, font=\footnotesize\bfseries] at (1222,34) {Round 2: {\color{bpOurs}\method}};
      \node[font=\scriptsize, text width=301\figpx] at (323.5,128) {What trend does the left plot show?};
      \node[font=\scriptsize, text width=349\figpx] at (867.5,128) {What is the lowest y tick of the right plot?};
      \node[font=\scriptsize, text width=349\figpx] at (1543.5,128) {What is the lowest y tick of the right plot?};
      \node[font=\footnotesize] at (316,451.5) {long reasoning};
      \foreach \x in {96, 319} {\node[anchor=east, font=\scriptsize] at (\x,240) {0.4}; \node[anchor=east, font=\scriptsize] at (\x,305) {0.2};}
      \node[font=\scriptsize] at (712.5,228) {deleted during};
      \node[font=\scriptsize] at (712.5,270) {reasoning};
      \node[font=\scriptsize\bfseries, color=bpBase] at (920,270) {kept};
      \node[anchor=east, font=\scriptsize] at (1088,480) {0.4};
      \node[anchor=east, font=\scriptsize] at (1088,560) {0.2};
      \node[font=\footnotesize] at (866.5,729.5) {0.4};
      \node[font=\scriptsize] at (1350,258) {\textbf{Shelf} (CPU)};
      \node[font=\scriptsize] at (1350,302) {never deleted};
      \node[font=\scriptsize] at (1768,258) {\textbf{Desk} (GPU)};
      \node[font=\scriptsize] at (1768,302) {read by attention};
      \node[font=\scriptsize] at (1559,550) {read on};
      \node[font=\scriptsize] at (1559,594) {demand};
      \node[font=\scriptsize] at (1839,532) {parked};
      \node[font=\footnotesize] at (1578.5,728.5) {0.2};
    \end{tikzpicture}
  \end{minipage}\hfill
  \begin{minipage}[b]{0.285\textwidth}
    \centering
    \begin{tikzpicture}
      \begin{axis}[
        ybar, width=1.12\linewidth, height=4.1cm, ymin=40, ymax=100, enlarge x limits=0.75, bar width=7pt, axis lines*=left,
        ylabel={Accuracy (\%)}, ylabel style={font=\scriptsize}, ticklabel style={font=\scriptsize},
        title={(b) Same question and GPU memory}, title style={font=\footnotesize, at={(0.45,1)}},
        symbolic x coords={first, follow}, xtick=data, xticklabels={{Asked first\\{\color{bpBase}\textminus 8.3 vs.\ full}}, {After reasoning\\{\color{bpBase}\textminus 24.3 vs.\ full}}}, xticklabel style={align=center},
        ytick={40,50,60,70,80,90}, ymajorgrids, grid style={gray!20},
        nodes near coords, every node near coord/.append style={font=\scriptsize, color=black, rotate=90, anchor=west},
        nodes near coords style={/pgf/number format/fixed, /pgf/number format/precision=1, /pgf/number format/fixed zerofill},
        legend style={font=\scriptsize, draw=none, fill=none, legend columns=1, at={(0.5,-0.36)}, anchor=north, row sep=-1pt}, legend cell align=left,
        legend image code/.code={\draw[#1, draw=none] (0cm,-0.06cm) rectangle (0.22cm,0.1cm);},
        clip=false,
      ]
        \addplot[fill=bpBase, draw=none] coordinates {(first,79.3) (follow,61.3)};
        \addplot[fill=bpOurs, draw=none] coordinates {(first,85.3) (follow,81.7)};
        \addplot[fill=bpText!60, draw=none] coordinates {(first,87.7) (follow,85.7)};
        \legend{DBTrimKV, \method on DBTrimKV, Full cache}
      \end{axis}
    \end{tikzpicture}
  \end{minipage}
  \caption{\textbf{Keep the reasoning, read the image.}
  (a) A retention policy decides what to keep while the model reasons, so a follow-up that needs a region the reasoning did not use finds it deleted.
  \method keeps every visual entry on a shelf in CPU memory and moves the entries the current query needs onto the desk that attention reads.
  (b) Under DBTrimKV, a descriptive CharXiv question loses three times as much accuracy after a round of reasoning as when asked first; reading the image on demand at matched GPU memory closes most of the gap.}
  \label{fig:teaser}
\end{center}

\vspace{0.6em}
}]
{\renewcommand{\thefootnote}{*}\footnotetext{Equal contribution. Names are listed in alphabetical order by surname.}}

\begin{abstract}
Reasoning vision-language models (VLMs) think for thousands of tokens before they answer, and retention policies keep the KV cache affordable by pruning it as the reasoning grows.
We find that this pruning sacrifices the image in multi-turn conversations: the image and the reasoning share one budget, so the growing reasoning crowds the visual entries out.
A chart question that loses 8 points to the full cache when asked first loses 24 points when asked after a round of reasoning.
Protecting a small visual reserve from eviction recovers most of this loss, while adding the same slots to the shared budget recovers considerably less.
The image therefore lacks protection, not capacity.
Based on this finding, we propose \method, a training-free method that works on top of existing retention policies.
It keeps every visual entry in CPU memory and, at each decoding step, swaps the visual entries that the current query scores highest into the attention slots of the text entries it scores lowest.
The image is never evicted, yet it occupies attention slots only while it is read.
We evaluate \method on four benchmarks, three of them conversational, with two retention policies and two reasoning VLMs.
In conversations, it improves the strongest existing policy, DBTrimKV, by up to 20.3 points at matched GPU memory, and still leads by 8.7 points when DBTrimKV uses twice the memory.
It also consistently outperforms a fixed visual reserve of matched GPU memory in these settings, without hurting accuracy on standalone questions.
These results show that the image should be protected from the reasoning but read only on demand.
\end{abstract}

\section{Introduction}
\label{sec:intro}

Reasoning vision-language models (VLMs) such as Qwen3-VL-Thinking~\cite{qwen3vl} solve chart, diagram and mathematics problems by thinking for thousands of tokens before they answer.
Each token adds an entry to the key-value (KV) cache, so the cost of attention grows with the reasoning.
Retention policies keep this cost affordable by holding a fixed number of entries per head and deleting the rest.
Heuristic policies such as H2O~\cite{h2o}, SnapKV~\cite{snapkv} and R-KV~\cite{rkv} rank entries by attention and redundancy, while DBTrimKV~\cite{dbtrimkv} learns a retention gate for every head and preserves accuracy at budgets where the heuristics collapse.

These policies let the image and the reasoning compete for one budget, so both are kept or deleted by the same rule.
A conversation exposes this weakness. In \cref{fig:teaser}(a), after the model has reasoned at length about a chart, the user asks about a region that the reasoning never used.
The attention of earlier turns is already a poor guide to the regions that later turns need~\cite{seensaid}, and with long reasoning in between, the image must also compete with a growing amount of text.
Prior work already treats the two modalities differently during eviction~\cite{asymvlm,madakv}, yet the image still holds a fixed place in the cache or loses it for good.
Methods that read the cache on demand keep every entry~\cite{quest,shadowkv,omnikv}, but they let the reasoning compete with the image for the same attention slots.

We find that in such conversations the reasoning crowds out the image.
As \cref{fig:teaser}(b) shows, on follow-up questions built from CharXiv~\cite{charxiv}, DBTrimKV with 256 entries per head loses 8.3 points to the full cache when a question comes first, but 24.3 points when the same question follows a round of reasoning.
The image and the question are identical in both cases, so the extra loss comes from the reasoning in between: by the time the follow-up arrives, DBTrimKV has deleted 97.9\% of the image's visual entries.
To test whether the image needs more room or only protection, we call the retention policy the \emph{host}, let it manage only the text, and protect a small \emph{visual reserve} that it selects once the prompt has been read.
A reserve of 128 entries per head recovers 17.7 points, which is 10.3 points more than the same entries recover when the host may evict them.
The image therefore lacks protection, not capacity.

A fixed reserve, however, takes its slots from the reasoning at every step, including the many steps at which the text needs them more.
We therefore propose \method, a training-free method that runs on top of an existing retention policy.
\method keeps every visual entry in CPU memory while the host manages the text on the GPU.
As \cref{fig:teaser}(a) shows, at each decoding step it swaps the visual entries the current query scores highest into the attention slots of the lowest-scoring text entries.
A compact sketch of the visual keys proposes candidates for the swap, so only a few keys are read in full.
The cache keeps the reasoning and reads the image: the image is never evicted, yet it occupies attention slots only while a query reads it.

We evaluate \method on three conversational benchmarks and one of standalone questions, with DBTrimKV and R-KV as hosts and two Qwen3-VL-Thinking models.
On the CharXiv follow-ups, \method with 259 GPU entries per head improves DBTrimKV by 20.3 points at matched GPU memory and still leads by 8.7 points when DBTrimKV uses twice the memory; with 386 entries, it comes within 1.7 points of the full cache.
It also outperforms the best fixed split between text and image by up to 12.7 points.
Against a fixed visual reserve of matched GPU memory, it stays ahead in five-turn conversations, on the multi-image MMDU dialogues, with R-KV as host and with the second model, without hurting accuracy on standalone questions.

Our contributions are threefold:
\begin{itemize}
    \item We show that KV retention sacrifices the image in conversations with long reasoning, and that the cause is missing protection, not missing capacity.
    \item We propose \method, a training-free method that keeps the image in CPU memory and reads it on demand.
    \item We validate \method on four benchmarks with two retention policies and two models, where it consistently outperforms a fixed visual reserve of matched GPU memory in conversations.
\end{itemize}
\section{Related Work}
\label{sec:related}

\paragraph{KV cache retention.}
Retention policies bound the KV cache by deleting entries during generation.
Early policies such as H2O~\cite{h2o}, StreamingLLM~\cite{streamingllm} and SnapKV~\cite{snapkv} select entries by attention and recency, and Ada-KV~\cite{adakv} divides the budget across heads.
For reasoning models, R-KV~\cite{rkv} also removes redundant tokens and EpiKV~\cite{epikv} avoids the attention matrix, while recent studies report that eviction at low budgets can lengthen the reasoning~\cite{kvreasoning} and that random eviction outside the prompt can match the best selectors~\cite{randomattn}.
Learned policies such as TrimKV~\cite{trimkv} and ForesightKV~\cite{foresightkv} train the eviction rule instead.
Among them, DBTrimKV~\cite{dbtrimkv} lets entries from all layers, heads and modalities compete for one global budget, and it is the strongest policy in our experiments.
All of them delete what they evict, so a visual entry removed during reasoning is gone when a later question needs it.
\method keeps such a policy as the host for the text and takes the image out of its budget.

\paragraph{Visual compression in VLMs.}
Most work on efficient VLM inference compresses the image.
Token pruning methods such as FastV~\cite{fastv}, VisionZip~\cite{visionzip} and DivPrune~\cite{divprune} remove image tokens at prefill according to attention or feature diversity~\cite{pyramiddrop,sparsevlm}, and MetaCompress~\cite{metacompress} learns a reduction that does not depend on the question.
KV compression methods prune or merge the prompt cache once, as in LOOK-M~\cite{lookm}, MEDA~\cite{meda} and FlashCache~\cite{flashcache}, or make the budget depend on the modality, as in VL-Cache~\cite{vlcache}, MadaKV~\cite{madakv}, AsymVLM~\cite{asymvlm} and TGV-KV~\cite{tgvkv}.
These methods decide which visual entries survive before the reasoning starts, yet the regions a model needs change during decoding and across turns~\cite{seensaid,dstp}.
A few methods revise the choice later.
DSTP~\cite{dstp}, one of our baselines, adds tokens back when attention drifts, DyCoke~\cite{dycoke} and \mbox{SparseVILA}~\cite{sparsevila} select again at every step or for every new question, and RetentiveKV~\cite{retentivekv} folds evicted entries into a compact memory.
They still leave the reasoning to grow or keep only an approximation of what they evict.
\method keeps every visual entry exact, bounds the reasoning with the host, and gives the image attention slots only when a query needs it.

\paragraph{Reading the cache on demand.}
Another line of work keeps every entry and reads only the ones that the current query needs.
Quest~\cite{quest}, Loki~\cite{loki} and SparQ~\cite{sparq} find these entries with cheap approximate scores, while ArkVale~\cite{arkvale}, OmniKV~\cite{omnikv}, InfiniGen~\cite{infinigen} and ShadowKV~\cite{shadowkv} store part of the cache in CPU memory and fetch it when needed.
\mbox{QEvict}~\cite{qevict} further moves rarely attended text to a quantized tier.
None of them separates the image from the reasoning, so the two compete for the same attention slots.
For VLMs, several studies find that attention to the image fades as the reasoning grows~\cite{qwenlookagain,lookback,tvc,pointvisual}.
They restore it by training the model to look again~\cite{qwenlookagain,lookback,reground} or by feeding image tokens back into the model~\cite{qwenlookagain,tvc,pointvisual,visref,memvr}.
Concurrent with our work, ViMoD~\cite{vimod} trains a router that recalls visual KV during reasoning, and RCR~\cite{rcr} selects its visual KV again when attention drifts.
These methods keep the full cache, lengthen the context or need training, whereas \method reads only the image on demand, under a fixed budget per head and without training.
\section{The Reasoning Crowds Out the Image}
\label{sec:finding}

After a round of reasoning, the image is unchanged, but what the model can still read from its cache may not be.
Under a fixed KV budget, generated text keeps entering the cache while visual entries may be evicted.
We ask whether this competition harms follow-up answers, and whether the image needs more room or protection from the reasoning.

\subsection{Preliminaries}
\label{sec:prelim}

We call the underlying retention policy the \emph{host} and use DBTrimKV~\cite{dbtrimkv} as our main host.
The cache holds visual entries $\V$, prompt text $\A$ (the system prompt and the first question), and text $\R$ added after the prompt (the reasoning and later turns).
The host keeps a set $H_t$ at step $t$ under a budget $B$ and deletes the rest for good.
DBTrimKV pools this budget over all layers and KV heads, so $B$ is the average number of entries per head.

We use Qwen3-VL-8B-Thinking~\cite{qwen3vl} on the \emph{CharXiv follow-ups}, two-turn conversations built from CharXiv~\cite{charxiv}.
The first turn asks a reasoning question about a chart; the second asks a descriptive question about the same chart.
To keep the first turn identical across methods, we generate its reasoning and answer with the full-cache model and feed them token by token through each method's cache.
Every method thus sees the same reasoning but keeps it with its own policy, and we score the follow-up answer.

\subsection{The image is evicted during reasoning}
\label{sec:crowd}

\cref{fig:teaser}(b) shows that preceding reasoning widens the accuracy gap caused by cache compression.
When a descriptive question is asked directly, DBTrimKV at $B = 256$ loses 8.3 points to the full cache.
When the same question follows a round of reasoning, the gap grows to 24.3 points.

\begin{figure}[t]
  \centering
  \input{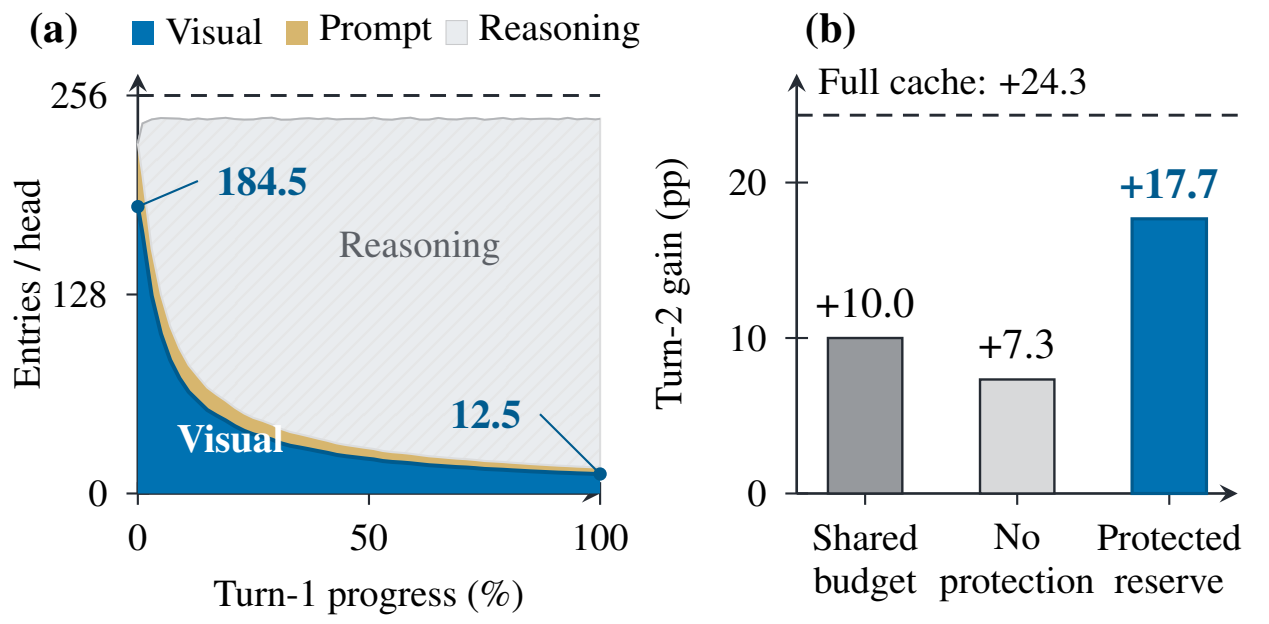}
  \caption{
    \textbf{Reasoning crowds out visual KV; protection beats extra shared capacity.}
    (a) Cache composition during the first-turn output under DBTrimKV ($B = 256$).
    (b) Follow-up accuracy gains (percentage points) over the same host (61.3\%); the dashed line marks the unconstrained full-cache gain.
  }
  \label{fig:finding}
\end{figure}

\cref{fig:finding}(a) shows what happens inside the cache.
From the end of prefill to the end of the first-turn output, the average number of retained visual entries per KV head falls from 184.5 (already pruned at prefill) to 12.5, while generated text comes to occupy most of the cache.
The host ranks the image and the text with one rule under one shared budget.
Every new token adds a competitor, and once a visual entry is evicted, the host cannot read it again for a follow-up.
By the time the follow-up arrives, the host has deleted 97.9\% of the visual entries.

This exposes a mismatch in multi-turn reasoning: visual information that the current reasoning no longer needs may still be needed by a later question.
Yet the host must decide which visual entries to delete before that question is asked.

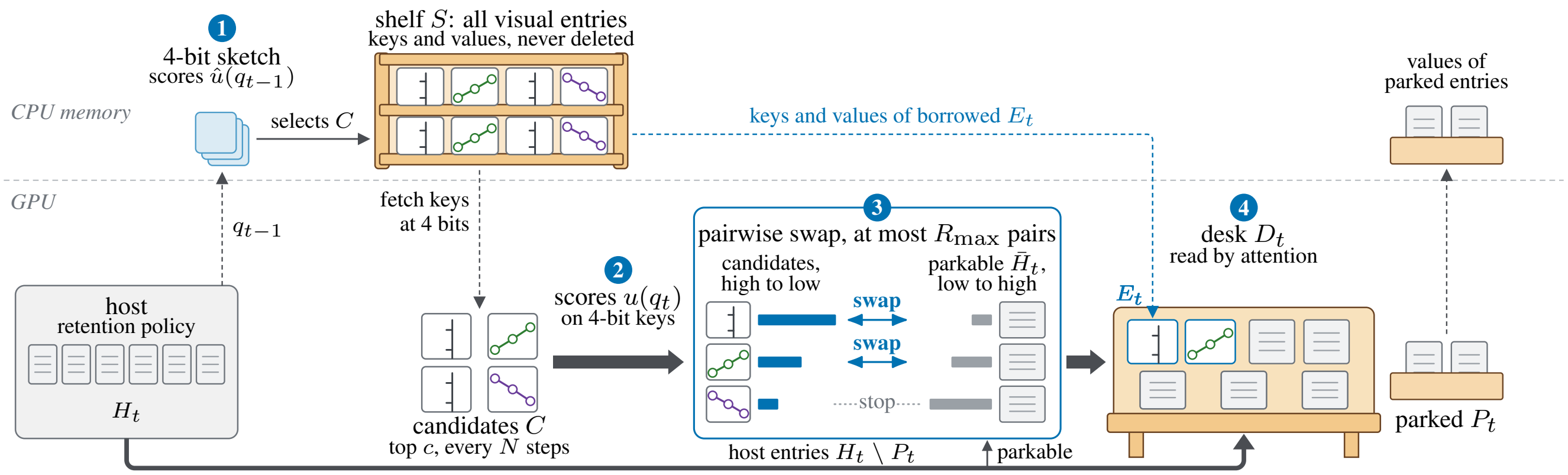
\begin{figure*}[!t]
  \centering
  \setlength{\figpx}{0.000362\textwidth}%
  \begin{tikzpicture}[x=1.364\figpx, y=-1.364\figpx, every node/.style={inner sep=0pt, align=center}]
    \newcommand{\fStep}[3]{\filldraw[fill=bpOurs, draw=bpOurs] (#1,#2) circle (17); \node[font=\scriptsize\bfseries, text=white] at (#1,#2) {#3};}
    \newcommand{\fArr}[3]{\fill[fArrow] (#1,#3-8) rectangle (#2-18,#3+8); \fill[fArrow] (#2-20,#3-16) -- (#2,#3) -- (#2-20,#3+16) -- cycle;}
    \draw[fDash!70, line width=2.2\figpx, dash pattern=on 10\figpx off 7\figpx] (0,235) -- (2000,235);
    \node[anchor=west, font=\scriptsize\itshape, text=fTile] at (8,150) {CPU memory};
    \node[anchor=west, font=\scriptsize\itshape, text=fTile] at (8,262) {GPU};
    \node[font=\footnotesize] at (638,28) {shelf $S$: all visual entries};
    \node[font=\scriptsize] at (638,56) {keys and values, never deleted};
    \filldraw[fill=fWood, draw=fWoodD, line width=2.6\figpx, rounded corners=6\figpx] (476,72) rectangle (494,220);
    \filldraw[fill=fWood, draw=fWoodD, line width=2.6\figpx, rounded corners=6\figpx] (782,72) rectangle (800,220);
    \filldraw[fill=fWood!45, draw=fWoodD, line width=2.2\figpx] (494,84) rectangle (782,206);
    \foreach \y in {80,145,208} \filldraw[fill=fWood, draw=fWoodD, line width=2.6\figpx, rounded corners=3\figpx] (482,\y-7) rectangle (794,\y+7);
    \foreach \r/\ya/\yb in {0/91/139, 1/154/202} \foreach \c/\k in {0/1,1/2,2/1,3/3} {\pgfmathsetmacro{\xx}{504+\c*70}\fTile{\xx}{\ya}{\xx+60}{\yb}{\k}}
    \node[font=\scriptsize] at (1850,82) {values of};
    \node[font=\scriptsize] at (1850,110) {parked entries};
    \fCard{1798}{140}{1844}{196}
    \fCard{1856}{140}{1902}{196}
    \filldraw[fill=fWood!70, draw=fWoodD, line width=2.4\figpx, rounded corners=4\figpx] (1778,180) rectangle (1922,214);
    \begin{scope}[yshift=81.84\figpx]%
    \filldraw[fill=fLabel!60, draw=fTile, line width=2.6\figpx, rounded corners=16\figpx] (15,430) rectangle (300,626);
    \node[font=\footnotesize] at (157.5,456) {host};
    \node[font=\scriptsize] at (157.5,484) {retention policy};
    \foreach \i in {0,...,5} {\pgfmathsetmacro{\xx}{32+\i*43}\fCard{\xx}{506}{\xx+36}{556}}
    \node[font=\scriptsize] at (157.5,592) {$H_t$};
    \end{scope}
    \fStep{280}{40}{1}
    \node[font=\footnotesize] at (280,76) {4-bit sketch};
    \node[font=\scriptsize] at (280,104) {scores $\hat u(q_{t-1})$};
    \foreach \d in {16,8,0} \filldraw[fill=bpOurs!14, draw=bpOurs!70, line width=2.4\figpx, rounded corners=8\figpx] (246+\d,148+\d) rectangle (298+\d,200+\d);
    \draw[fArrow, line width=2.4\figpx, dash pattern=on 8\figpx off 6\figpx] (280,370) -- (280,238);
    \fill[fArrow] (272,240) -- (280,224) -- (288,240) -- cycle;
    \node[anchor=west, font=\scriptsize] at (292,300) {$q_{t-1}$};
    \draw[fArrow, line width=3\figpx] (324,182) -- (456,182);
    \fill[fArrow] (454,174) -- (470,182) -- (454,190) -- cycle;
    \node[font=\scriptsize] at (396,158) {selects $C$};
    \begin{scope}[yshift=81.84\figpx]%
    \fTile{536}{466}{599}{524}{1}
    \fTile{621}{466}{684}{524}{2}
    \fTile{536}{534}{599}{592}{1}
    \fTile{621}{534}{684}{592}{3}
    \end{scope}
    \draw[fArrow, line width=2.4\figpx, dash pattern=on 8\figpx off 6\figpx] (610,222) -- (610,386);
    \fill[fArrow] (602,384) -- (610,400) -- (618,384) -- cycle;
    \node[anchor=east, font=\scriptsize] at (598,262) {fetch keys};
    \node[anchor=east, font=\scriptsize] at (598,290) {at 4 bits};
    \begin{scope}[yshift=81.84\figpx]%
    \node[font=\footnotesize] at (610,609) {candidates $C$};
    \node[font=\scriptsize] at (610,639) {top $c$, every $N$ steps};
    \end{scope}
    \begin{scope}[yshift=81.84\figpx]%
    \fStep{788}{410}{2}
    \node[font=\footnotesize] at (788,446) {scores $u(q_t)$};
    \node[font=\scriptsize] at (788,474) {on 4-bit keys};
    \fArr{704}{872}{528}
    \filldraw[fill=white, draw=bpOurs, line width=3.2\figpx, rounded corners=18\figpx] (885,330) rectangle (1355,626);
    \fStep{1120}{330}{3}
    \node[font=\footnotesize] at (1120,366) {pairwise swap, at most $R_{\max}$ pairs};
    \node[font=\scriptsize] at (984,402) {candidates,};
    \node[font=\scriptsize] at (984,430) {high to low};
    \node[font=\scriptsize] at (1261,402) {parkable $\bar H_t$,};
    \node[font=\scriptsize] at (1261,430) {low to high};
    \foreach \y/\k/\lv/\lt in {474/1/100/26, 528/2/56/52, 582/3/26/80} {
      \fTile{901}{\y-23}{957}{\y+23}{\k}
      \fill[bpOurs, rounded corners=2\figpx] (967,\y-7) rectangle (967+\lv,\y+7);
      \fill[fDash, rounded corners=2\figpx] (1267-\lt,\y-7) rectangle (1267,\y+7);
      \fCard{1277}{\y-23}{1335}{\y+23}
    }
    \foreach \y in {474,528} {
      \draw[bpOurs, line width=3.4\figpx, {Latex[round]}-{Latex[round]}] (1082,\y) -- (1158,\y);
      \node[font=\scriptsize\bfseries, text=bpOurs] at (1120,\y-20) {swap};
    }
    \draw[fDash, line width=3\figpx, dash pattern=on 2\figpx off 7\figpx, line cap=round] (1065,582) -- (1094,582) (1146,582) -- (1175,582);
    \node[font=\scriptsize, text=fTile] at (1120,582) {stop};
    \fArr{1363}{1413}{528}
    \fStep{1590}{330}{4}
    \node[font=\footnotesize] at (1590,366) {desk $D_t$};
    \node[font=\scriptsize] at (1590,394) {read by attention};
    \filldraw[fill=fWood, draw=fWoodD, line width=2.6\figpx] (1443,612) rectangle (1461,650) (1719,612) rectangle (1737,650);
    \filldraw[fill=fDesk, draw=fWoodD, line width=2.6\figpx, rounded corners=14\figpx] (1423,458) rectangle (1757,604);
    \filldraw[fill=fWood, draw=fWoodD, line width=2.6\figpx, rounded corners=5\figpx] (1415,594) rectangle (1765,618);
    \fTile[bpOurs]{1441}{474}{1505}{530}{1}
    \fTile[bpOurs]{1515}{474}{1579}{530}{2}
    \foreach \x in {1597,1667} \fCard{\x}{474}{\x+58}{530}
    \foreach \x in {1457,1561,1665} \fCard{\x}{540}{\x+58}{588}
    \node[anchor=east, font=\scriptsize, text=bpOurs] at (1463,442) {$E_t$};
    \fCard{1798}{502}{1844}{558}
    \fCard{1856}{502}{1902}{558}
    \filldraw[fill=fWood!70, draw=fWoodD, line width=2.4\figpx, rounded corners=4\figpx] (1778,542) rectangle (1922,576);
    \node[font=\footnotesize] at (1850,604) {parked $P_t$};
    \end{scope}
    \draw[fArrow, line width=2.4\figpx, dash pattern=on 8\figpx off 6\figpx] (1850,432) -- (1850,232);
    \fill[fArrow] (1842,236) -- (1850,220) -- (1858,236) -- cycle;
    \draw[bpOurs, line width=2.6\figpx, dash pattern=on 8\figpx off 6\figpx, rounded corners=10\figpx] (806,176) -- (1473,176) -- (1473,398);
    \fill[bpOurs] (1465,396) -- (1473,412) -- (1481,396) -- cycle;
    \node[font=\scriptsize, text=bpOurs] at (1139.5,154) {keys and values of borrowed $E_t$};
    \begin{scope}[yshift=81.84\figpx]%
    \draw[fArrow, line width=8\figpx, rounded corners=14\figpx] (157.5,626) -- (157.5,664) -- (1590,664) -- (1590,638);
    \fill[fArrow] (1576,640) -- (1590,620) -- (1604,640) -- cycle;
    \draw[fArrow, line width=3\figpx] (1261,664) -- (1261,640);
    \fill[fArrow] (1254,642) -- (1261,628) -- (1268,642) -- cycle;
    \node[anchor=west, font=\scriptsize] at (1273,645) {parkable};
    \node[font=\scriptsize] at (1050,645) {host entries $H_t \setminus P_t$};
    \end{scope}
  \end{tikzpicture}
\caption{
  \textbf{\method: keep the image, read it on demand.}
  Every visual entry stays on the shelf in CPU memory.
  A sketch proposes candidates, and the current query swaps those that outscore text entries onto the desk on the GPU.
  Displaced text entries are parked, not deleted, so the desk keeps the host's entry count.
}
\label{fig:overview}

\end{figure*}

\subsection{Protection, not capacity}
\label{sec:reserve}

Is the image evicted simply because the cache is too small?
We compare three configurations of the same total size (\cref{fig:finding}b): a \emph{shared budget} of 384 instead of 256; a \emph{protected reserve} of 128 visual entries per head on average, with the host managing the text under a budget of 256; and the same initial cache with \emph{no protection}, so the host may still evict its visual entries.
The reserve selects entries with the host's own retention score.
The latter two configurations start with exactly the same visual entries and differ only in whether those entries remain protected.

As \cref{fig:finding}(b) shows, enlarging the shared budget raises follow-up accuracy from 61.3\% to 71.3\%.
The same initial visual entries without protection reach 68.7\%, while the protected reserve reaches 79.0\%.
The reserve thus recovers 17.7 of the 24.3 points lost to the full cache, against 10.0 for the larger shared budget and 7.3 for the same entries left unprotected.
Extra space helps, but protection yields a larger improvement at the same capacity.
The unprotected control further shows that selecting visual entries at the start is not enough: they must survive the reasoning.

\section{ShelfKV: Reading the Image on Demand}
\label{sec:method}

Visual entries need to survive the reasoning, but need not occupy attention slots throughout.
\method therefore separates keeping visual KV from reading it: the host manages the text, every visual entry stays in CPU memory, and when the current query needs the image, visual entries temporarily borrow attention slots from the text.
Borrowing deletes no text and does not enlarge the set that attention reads.

\cref{fig:overview} shows how \method works.
Every few steps, a compact sketch selects candidate visual entries.
At every step, the current query compares these candidates with the host's text entries, higher-scoring visual entries take the slots of lower-scoring text entries, and attention reads the resulting set.
Candidate selection cuts the search cost; the swap sets how many slots the image takes.

\subsection{The shelf and the desk}
\label{sec:shelf}

The host manages only the text $\A \cup \R$ and spends its budget on text retention, while every visual entry stays on the \emph{\shelf} $S = \V$ in CPU memory throughout the conversation.
We call the set of entries a KV head reads at step $t$ the \emph{\desk}:
\begin{gather}
  D_t = (H_t \setminus P_t) \cup E_t, \nonumber\\
  E_t \subseteq S,\quad P_t \subseteq H_t,\quad
  |E_t| = |P_t| \le R_{\max}.
  \label{eq:desk}
\end{gather}
Here, $E_t$ contains the borrowed visual entries, $P_t$ the text entries temporarily parked to make room, and $R_{\max}$ caps the number of borrowed slots.

Each borrowed visual entry takes the slot of one parked text entry, so $|D_t| = |H_t|$.
Parking is not eviction: parked entries remain in the host's set and return to the desk when no longer displaced.
The host decides which text entries to keep or evict, and \method decides which entries attention reads at each step: an entry's lifetime and its use are managed separately.

\subsection{Pairwise swap}
\label{sec:swap}

A visual entry borrows a desk slot only when the current query scores it above the text entry it would displace, so \method scores visual candidates and text on one scale.
For $G$ query heads sharing a KV head, we aggregate their attention logits by log-sum-exp:
\begin{equation}
  u_i = \log \sum_{g=1}^{G}
  \exp\!\left(\frac{q^{(g)} \cdot k_i}{\sqrt{d}}\right),
  \label{eq:score}
\end{equation}
where $d$ is the head dimension.
For a single query head, this score orders entries as the attention weights do; for grouped queries, it combines relevance across the query heads into one score.

To preserve the current text context, the host's retained prompt text and the most recent $W$ generated tokens always stay on the desk, so the parkable set is
\begin{equation}
  \bar H_t = H_t \setminus (\A \cup \R_{t-W:t}).
  \label{eq:parkable}
\end{equation}
We sort the visual candidates $C_t$ (\cref{eq:candidates}) by decreasing score,
$u_{s_1} \ge u_{s_2} \ge \cdots$,
and parkable text entries by increasing score,
$u_{p_1} \le u_{p_2} \le \cdots$,
and compare them in pairs, starting with the highest-scoring visual entry and the lowest-scoring text entry.
We swap each pair while the visual score exceeds the text score by more than $\delta$, stopping at the first failed comparison or after $R_{\max}$ swaps.
With $m_t = \min\{R_{\max}, |C_t|, |\bar H_t|\}$ candidate pairs, the number of swaps is
\begin{equation}
  n_t = \sum_{j=1}^{m_t}
  \mathbf{1}\!\left[u_{s_j} > u_{p_j} + \delta\right].
  \label{eq:swap}
\end{equation}

We use $\delta = 0$, so each swap gives a slot to a higher-scoring entry.
The rule is recomputed at every step and layer for each KV head, with no swap when no candidate wins.
The image thus takes no fixed share of the desk: it borrows slots according to its score relative to the text under the current query, up to $R_{\max}$.

\subsection{Sketch candidates}
\label{sec:sketch}

Pairwise swaps decide which visual entries enter attention, but reading every visual key at every step to find candidates would defeat the purpose of reading on demand.
\method therefore searches the shelf with a compact sketch and leaves the current query to decide which candidates are read.
At the first decoding step, we fit a principal component basis to the visual keys of each sample, layer and KV head, obtaining the mean $\mu$ and a projection matrix $U$ of the top $r$ principal directions.
We project each visual key, $z_i = U(k_i - \mu)$, and quantize it per dimension to $b$ bits, giving its sketch $\hat z_i$.
Projecting the query into the same space lets us estimate relevance with the scoring rule of the swap:
\begin{equation}
  \hat u_i(q) =
  \log \sum_{g=1}^{G} \exp\!\left(
  \frac{(Uq^{(g)})^\top \hat z_i + (q^{(g)})^\top \mu}{\sqrt{d}}
  \right).
  \label{eq:sketch}
\end{equation}
Query projections and mean terms are computed once per query, so each key is scored in the sketch space without transferring or reconstructing full-precision keys.

Every $N$ steps, the sketch refreshes the candidate set with the $c$ visual entries that score highest under the previous query, so that selection can overlap the current step; the set is kept until the next refresh:
\begin{equation}
  C_t = \operatorname{Top\text{-}c}_{i \in S}
  \hat u_i(q_{t-1}).
  \label{eq:candidates}
\end{equation}
Their keys are transferred to the GPU at 4 bits, overlapping the current step's computation, and reused between refreshes.
At every step, the current query scores these keys and applies the pairwise swap, and only the selected visual entries are read in full precision.

\begin{table*}[!t]
  \centering
  \begin{minipage}{0.92\textwidth}
    \centering
    \small
    \setlength{\tabcolsep}{3pt}
    \renewcommand{\arraystretch}{1.03}

    \begin{tabularx}{\linewidth}{@{}l*{6}{>{\centering\arraybackslash}X}@{}}
      \toprule
      \rowcolor{sktHeader}
      \textbf{Method}
      & \shortstack{\textbf{Host budget}\\$B$}
      & \shortstack{\textbf{GPU cache}\\entries/head}
      & \shortstack{\textbf{CharXiv}\\follow-ups\\Acc. (\%) $\uparrow$}
      & \shortstack{\textbf{CharXiv}\\conversations\\Acc. (\%) $\uparrow$}
      & \shortstack{\textbf{MMMU-Pro}\\Acc. (\%) $\uparrow$}
      & \shortstack{\textbf{MMDU}\\Score $\uparrow$} \\
      \midrule

      Full cache
      & -- & all & 85.7 & 87.3 & 54.0 & 5.2 \\
      DSTP + R-KV
      & 512 & 512 & 15.3 & 16.8 & 6.3 & 3.6 \\
      DSTP + R-KV
      & 1024 & 1024 & 48.0 & 49.2 & 21.3 & 4.1 \\

      \midrule
      \rowcolor{sktGroup}
      \multicolumn{7}{c}{
        \textcolor{sktAccent}{\textbf{Qwen3-VL-8B-Thinking, R-KV host}}
      } \\

      Host
      & 256 & 256 & 4.7 & 5.3 & 7.0 & 3.1 \\
      Host
      & 384 & 384 & 14.0 & 13.5 & 11.0 & 3.5 \\
      Host
      & 512 & 512 & 21.7 & 20.9 & 10.7 & 3.8 \\
      Host
      & 1024 & 1024 & 61.3 & 63.7 & 22.3 & 4.3 \\
      Host + reserve ($V=128$)
      & 256 & 384 & 17.3 & 17.8 & 11.0 & 3.9 \\
      \rowcolor{sktHighlight}
      \textcolor{sktAccent}{\textbf{Host + \method}}
      & 338$^\ddagger$ & 386$^\ddagger$
      & 42.7 & 41.9 & 27.7 & 4.4 \\

      \midrule
      \rowcolor{sktGroup}
      \multicolumn{7}{c}{
        \textcolor{sktAccent}{\textbf{Qwen3-VL-8B-Thinking, DBTrimKV host}}
      } \\

      Host
      & 128 & 128 & 50.7 & 51.0 & 34.7 & 3.9 \\
      Host
      & 256 & 256 & 61.3 & 62.7 & 47.7 & 4.2 \\
      Host
      & 384 & 384 & 71.3 & 69.7 & 54.0 & 4.3 \\
      Host
      & 512 & 512 & 73.0 & 72.1 & 50.7 & 4.5 \\
      Host
      & 1024 & 1024 & 78.0 & 79.4 & 51.7 & 4.5 \\
      Host + reserve ($V=128$)
      & 256 & 384 & 79.0 & 81.9 & 55.3 & 4.6 \\
      \rowcolor{sktHighlight}
      \textcolor{sktAccent}{\textbf{Host + \method}}
      & 338$^\ddagger$ & 386$^\ddagger$
      & \textbf{84.0} & \textbf{85.0}
      & \textbf{56.7} & \textbf{4.9} \\

      \midrule
      \rowcolor{sktGroup}
      \multicolumn{7}{c}{
        \textcolor{sktAccent}{\textbf{Qwen3-VL-4B-Thinking, R-KV host}}
      } \\

      Full cache
      & -- & all & 82.4 & 83.4 & 57.0 & 4.6 \\
      Host
      & 256 & 256 & 1.4 & 2.8 & 5.4 & 3.0 \\
      Host
      & 512 & 512 & 11.9 & 12.4 & 6.8 & 3.4 \\
      Host + reserve ($V=128$)
      & 256 & 384 & 9.8 & 10.0 & 8.2 & 3.8 \\
      \rowcolor{sktHighlight}
      \textcolor{sktAccent}{\textbf{Host + \method}}
      & 338$^\ddagger$ & 386$^\ddagger$
      & 24.1 & 25.6 & 18.3 & 4.0 \\

      \bottomrule
    \end{tabularx}

    \caption{\textbf{Main results.}
    Rows above the 4B block use Qwen3-VL-8B-Thinking; bold marks the best compressed 8B result.
    GPU cache: KV entries per head, including auxiliary data.
    CharXiv conversations score turns 2--4; MMDU reports the judge score (1--10).
    $^\ddagger$CharXiv values; 320/383 on MMMU-Pro and 296/383 on MMDU.}
    \label{tab:main}
  \end{minipage}
\end{table*}

\subsection{Memory cost and implementation}
\label{sec:memory}

\method reuses text slots through swaps, so the desk holds the same number of entries as the host, and additional GPU memory stores only the candidate keys and their quantization metadata.
Measured in full-precision KV entries, the host budget is $B$, the $c$ candidate keys at 4 bits cost $c/8$ entries, and the metadata costs one entry per group of 128 visual tokens.
For an image with $N_v$ visual tokens, the GPU cache cost is
\begin{equation}
  M_{\mathrm{GPU}}
  = B + \frac{c}{8}
  + \left\lceil \frac{N_v}{128} \right\rceil.
  \label{eq:memory}
\end{equation}

All visual KV and the sketch stay in CPU memory, and the GPU holds only the candidate keys and the current desk.
Selected full-precision visual data is read directly from pinned CPU memory through GPU gather kernels, so the full image remains available while attention slots serve the entries read at the current step.
\section{Experiments}
\label{sec:exp}

We evaluate \method on follow-up questions, multi-turn conversations and single-turn visual reasoning.
We compare it with the hosts and fixed visual reserves, analyze protection and memory allocation, and measure its decoding cost.

\subsection{Setup}
\label{sec:setup}

\paragraph{Models.}
We use Qwen3-VL-8B-Thinking~\cite{qwen3vl} as the main model and Qwen3-VL-4B-Thinking as a second model size, with generation settings and the evaluation framework shared across methods.
DBTrimKV uses its released pretrained gates; as it needs a trained gate per model, we test the 4B model with R-KV as the host.

\paragraph{Benchmarks.}
\emph{CharXiv follow-ups} pair a chart reasoning question with a descriptive question about the same chart~\cite{charxiv}; all methods receive the same first-turn output from the full-cache model, which isolates the effect of cache retention on the follow-up.
\emph{CharXiv conversations} chain the reasoning question and subsequent descriptive questions into a five-turn dialogue in which each method generates every answer itself, so they test whether the gains carry over to freely generated conversations.
\emph{MMMU-Pro Vision}~\cite{mmmupro} tests single-turn visual reasoning with questions printed in the image, and \emph{MMDU}~\cite{mmdu} tests open-ended answers in long, multi-image conversations.

\paragraph{Baselines.}
We use DBTrimKV~\cite{dbtrimkv} as the main host and R-KV~\cite{rkv} as a second host, and compare each host alone, with a fixed visual reserve, and with \method.
Additional baselines are the full cache and DSTP~\cite{dstp} combined with R-KV.

\paragraph{Metrics.}
We report accuracy on CharXiv and MMMU-Pro: CharXiv follow-ups score the second answer, and CharXiv conversations score the descriptive questions in turns 2 to 4, since the fifth cannot be answered from the chart.
Answers that hit the generation limit count as wrong, and point differences are computed before rounding.
MMDU answers receive model-judge scores from 1 to 10 using the official prompts.
Memory counts the GPU-resident cache and auxiliary data as in \cref{eq:memory}, not the nominal host budget alone, and runtime cost is measured by peak GPU memory and decoding time per step.

\paragraph{Protocol.}
We select \method's hyperparameters on development splits disjoint from the evaluation sets.
The image can borrow up to one quarter of the host's slots ($R_{\max}=B/4$), and the most recent $W=32$ generated tokens remain on the desk.
The sketch keeps $r=32$ dimensions at $b=4$ bits; every $N=4$ steps it selects $c=4R_{\max}$ candidates, whose keys the GPU holds at 4 bits.
The fixed visual reserve keeps $V=128$ entries per head on average, selected by the host's own retention score once the prompt has been read.
For \method, we lower the host budget to $B = 338$, 320 and 296 on CharXiv, MMMU-Pro and MMDU, so that its GPU cache in \cref{eq:memory} matches the reserve's 384 entries per head.
We call two configurations matched when their GPU caches differ by at most four entries per head.

\subsection{Main results}
\label{sec:main}

\cref{tab:main} shows that \method answers follow-ups more accurately with a smaller GPU cache.
On the CharXiv follow-ups, it reaches 84.0\% with 386 GPU entries per head and 81.7\% with 259 (\cref{tab:ablation}b), while DBTrimKV reaches 73.0\% at 512 entries and 78.0\% at 1024.
The gains do not depend on a fixed first turn: in freely generated CharXiv conversations and on MMDU, \method also outperforms DBTrimKV with 1024 entries per head (85.0\% vs.\ 79.4\%, and 4.9 vs.\ 4.5).

Compared with a fixed visual reserve at matched GPU memory, \method raises accuracy from 79.0\% to 84.0\% on the CharXiv follow-ups and from 81.9\% to 85.0\% on CharXiv conversations.
Since the reserve already prevents visual eviction, this gain comes from giving attention slots to the entries the current query needs.
The margin is smallest on single-turn MMMU-Pro (56.7\% vs.\ 55.3\%), where no later question needs regions that the reasoning did not use.

\paragraph{Other hosts and models.}
With R-KV, \method raises CharXiv follow-up accuracy from the reserve's 17.3\% to 42.7\% on the 8B model and from 9.8\% to 24.1\% on the 4B model, with consistent gains on the other benchmarks.
R-KV alone collapses at these budgets (4.7\% at $B=256$), which keeps all R-KV configurations well below the full cache.

\begin{figure}[t]
  \centering
  {\offinterlineskip
  \includegraphics[width=\columnwidth, trim=0 73.2bp 0 0, clip]{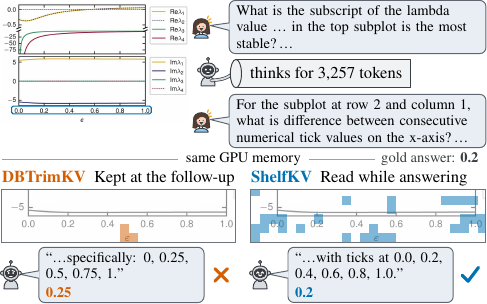}\par\vspace{3pt}
  \begin{tikzpicture}[x=0.0042830\columnwidth, y=-0.0042830\columnwidth]
    \node[anchor=north west, inner sep=0pt] at (0,0) {\includegraphics[width=\columnwidth, trim=0 28.1bp 0 72.25bp, clip]{figures/fig_case.pdf}};
    \fill[white] (230.3,1.95) rectangle (233.6,5.75);
  \end{tikzpicture}\par\vspace{3pt}
  \includegraphics[width=\columnwidth, trim=0 0 0 117.35bp, clip]{figures/fig_case.pdf}\par}
  \caption{\textbf{A follow-up that needs a region the reasoning never used.}
  The first turn asks about the top panel, and the follow-up asks for the x-axis ticks of the bottom panel (blue box).
  The crops show the visual entries that DBTrimKV still keeps when the follow-up begins and those that \method reads while answering, at the same GPU memory.
  Quotes are excerpts from the model outputs.}
  \label{fig:case}
\end{figure}

\paragraph{Qualitative example.}
\cref{fig:case} shows a CharXiv follow-up that DBTrimKV misses: the model reasons for 3,257 tokens about the top panel. When the follow-up asks for the x-axis ticks of the bottom panel, DBTrimKV keeps almost none of this region and answers with ticks that are not in the chart.
At the same GPU memory, \method reads the entries around the tick labels and reports them correctly.

\subsection{What does visual protection require?}
\label{sec:protect}

We fix DBTrimKV's text budget at $B=256$ and report in \cref{tab:reserve} how reserve size and entry selection affect visual protection.

\begin{table}[t]
  \centering
  \footnotesize
  \setlength{\tabcolsep}{3pt}
  \renewcommand{\arraystretch}{1.05}
  \begin{tabular}{@{}lccc@{}}
    \toprule
    Configuration & GPU & CharXiv & MMMU-Pro \\
    \midrule
    Host alone
    & 256 & 61.3 & 47.7 \\
    Shared budget ($B=384$)
    & 384 & 71.3 & 54.0 \\
    Unprotected ($V=128$)
    & 384 & 68.7 & 51.7 \\
    \rowcolor{sktHighlight}
    Reserve ($V=128$)
    & 384 & 79.0 & 55.3 \\
    \midrule
    Reserve ($V=32$)
    & 288 & 74.3 & 47.3 \\
    Reserve ($V=64$)
    & 320 & 81.0 & 53.0 \\
    Reserve ($V=256$)
    & 512 & 80.3 & 53.0 \\
    \midrule
    Random, matched counts
    & 384 & 78.2 & 35.7 \\
    Lowest-scoring, same total
    & 384 & 53.3 & 11.3 \\
    \midrule
    Whole image in attention
    & $256+N_v$ & 81.3 & 55.3 \\
    \bottomrule
  \end{tabular}
  \caption{
    \textbf{A small host-selected reserve suffices.}
    Accuracy (\%) with DBTrimKV at $B=256$,
    except for the shared-budget control.
    GPU counts equivalent KV entries per head;
    $N_v$ is the visual token count.
    Random selection on CharXiv is averaged over three draws.
  }
  \label{tab:reserve}
\end{table}

\paragraph{How much to protect?}
A small reserve already recovers much of the performance.
Even 32 entries per head raise CharXiv accuracy from 61.3\% to 74.3\%, although MMMU-Pro stays at the host's level (47.3\% vs.\ 47.7\%).
Keeping 64 visual entries per head gives 81.0\% on CharXiv, and enlarging the reserve brings no consistent further gain.
Compared with keeping the whole image in attention under the same text budget, a 128-entry reserve is only 2.3 points lower on CharXiv and equally accurate on MMMU-Pro.
Fixed protection thus need not keep the full image on the GPU.

\paragraph{Which entries to protect?}
Reserve size alone does not determine its benefit.
With the same entry counts, random selection averages 78.2\% on CharXiv, close to the host-selected reserve's 79.0\%, but reaches only 35.7\% on MMMU-Pro against 55.3\%.
Keeping the entries the host scores lowest performs poorly on both tasks.
Selection therefore matters more on MMMU-Pro, whose questions are printed in the image.

\subsection{Reading the image on demand}
\label{sec:ondemand}

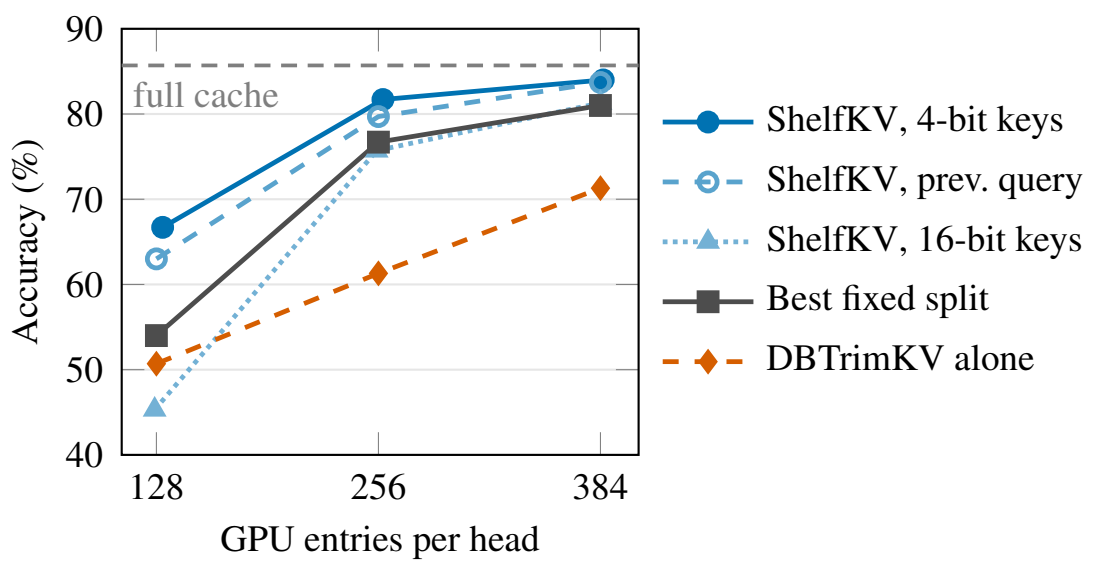
\begin{figure}[t]
  \centering
  \begin{tikzpicture}
    \begin{axis}[
      width=0.6\linewidth, height=4.4cm, scale only axis=false,
      xmin=108, xmax=406, ymin=40, ymax=90,
      xtick={128,256,384}, ytick={40,50,60,70,80,90},
      xlabel={GPU entries per head}, ylabel={Accuracy (\%)},
      label style={font=\scriptsize}, ticklabel style={font=\scriptsize},
      xlabel style={yshift=2pt}, ylabel style={yshift=-3pt},
      ymajorgrids, grid style={gray!20},
      legend style={font=\scriptsize, draw=none, fill=none, at={(1.04,0.5)}, anchor=west,
                    legend columns=1, row sep=1pt, inner sep=0pt},
      legend cell align=left,
      every axis plot/.append style={line width=0.9pt, mark size=1.7pt},
      clip=false,
    ]
      \addplot[bpOurs, mark=*] coordinates {(131.6,66.7) (258.6,81.7) (385.6,84.0)};
      \addlegendentry{\method, 4-bit keys}
      \addplot[bpOurs!65, dashed, mark=o, mark options={solid}] coordinates {(128,63.0) (256,79.7) (384,83.7)};
      \addlegendentry{\method, prev.\ query}
      \addplot[bpOurs!55, densely dotted, mark=triangle*, mark options={solid}] coordinates {(127,45.3) (255,75.7) (384,81.3)};
      \addlegendentry{\method, 16-bit keys}
      \addplot[black!70, mark=square*] coordinates {(128,54.0) (256,76.7) (384,81.0)};
      \addlegendentry{Best fixed split}
      \addplot[bpBase, dashed, mark=diamond*, mark options={solid}] coordinates {(128,50.7) (256,61.3) (384,71.3)};
      \addlegendentry{DBTrimKV alone}
      \addplot[bpText, densely dashed, line width=0.7pt, mark=none, forget plot] coordinates {(108,85.7) (406,85.7)};
      \node[font=\scriptsize, text=bpText, anchor=north west, inner sep=1pt] at (axis cs:111,84.6) {full cache};
    \end{axis}
  \end{tikzpicture}
\caption{
  \textbf{On-demand reading versus fixed allocation at matched GPU cache budgets.}
  CharXiv follow-ups with DBTrimKV.
  Best fixed split: the best text/visual allocation we ran that fits within each budget.
  16-bit keys: full-precision GPU candidate keys;
  prev.\ query: CPU candidates scored with the previous query.
}
\label{fig:ondemand}
\end{figure}

Once the image is protected, the space left to the text still matters, and a fixed reserve occupies its slots at every step.
\cref{fig:ondemand} compares \method with the best fixed split between text and image that fits within each GPU budget: at 128, 256 and 384 entries per head, \method leads by 12.7, 5.0 and 3.0 points.
At 384 entries, for example, the best split is the 64-entry reserve of \cref{tab:reserve} (81.0\% at 320 entries), against 84.0\% for \method.
The gap is largest at the tightest budget, where a fixed split leaves the text too few slots.
Which entries are read matters as well: reading the same number of visual entries at random reaches only 63.7\% at $B=256$, barely above the host alone (61.3\%).

\subsection{Ablations}
\label{sec:ablation}

\begin{table}[t]
  \centering
  \footnotesize
  \setlength{\tabcolsep}{3pt}
  \begin{tabular}{@{}lccc@{}}
    \toprule
    \multicolumn{4}{@{}l}{\textbf{(a) Mechanism}} \\
    \midrule
    Variant & \multicolumn{2}{c}{CharXiv} & MMMU-Pro \\
    \midrule
    \method, 16-bit keys (reference) & \multicolumn{2}{c}{81.3} & 52.7 \\
    No host (all entries on the shelf) & \multicolumn{2}{c}{80.0} & 47.3 \\
    Exact scores, no sketch & \multicolumn{2}{c}{81.3} & 51.3 \\
    \midrule
    \multicolumn{4}{@{}l}{\textbf{(b) Candidate-key storage}} \\
    \midrule
    Storage & \multicolumn{2}{c}{CharXiv} & MMMU-Pro \\
    GPU entries per head & 259 & 386 & 383 \\
    \midrule
    4-bit, GPU (\method) & 81.7 & 84.0 & 56.7 \\
    16-bit, GPU & 75.7 & 81.3 & 52.7 \\
    CPU, previous query & 79.7 & 83.7 & 53.7 \\
    \bottomrule
  \end{tabular}
  \caption{
    \textbf{Ablations.}
    Accuracy (\%) on the CharXiv follow-ups and MMMU-Pro.
    (a) Variants of \method with 16-bit candidate keys on DBTrimKV at $B=256$.
    (b) Rows in each column use matched GPU memory.
  }
  \label{tab:ablation}
\end{table}

\cref{tab:ablation} tests the components of \method: part (a) varies the mechanism, starting from \method with 16-bit candidate keys, and part (b) compares three ways to store the candidate keys at matched GPU memory.

\paragraph{The host should keep the reasoning.}
Without the host, the reasoning is also read from the shelf on demand, with exact scores for all entries.
MMMU-Pro then drops by 5.3 points, and the share of answers that hit the generation limit rises from 7.7\% to 19.0\%.
The CharXiv follow-ups are less affected (80.0\% vs.\ 81.3\%).
Reading on demand suits the image, whose size is fixed, but not the reasoning, which later steps keep using.

\paragraph{The sketch costs no accuracy.}
Scoring every visual key exactly at every step reads the whole shelf each time but gains nothing over the sketch: 81.3\% on CharXiv in both cases, and 51.3\% against 52.7\% on MMMU-Pro.
A sketch of $r=32$ dimensions at 4 bits is thus enough to propose the candidates.

\paragraph{Candidate keys stay on the GPU at 4 bits.}
At 259 entries per head, 16-bit candidate keys cut the host budget from 225 to 171 and cost 6.0 points. Candidates kept in CPU memory must be scored with the previous query and cost 2.0 points.
At 386 entries per head, CPU candidates come within 0.3 points of 4-bit keys on CharXiv, while 16-bit keys still trail by 2.7.
Keys on the GPU at 4 bits are the most accurate in every column.

\subsection{Cost}
\label{sec:cost}

\begin{table}[t]
  \centering
  \footnotesize
  \setlength{\tabcolsep}{2.6pt}
  \begin{tabular}{@{}lcccccccc@{}}
    \toprule
    & \multicolumn{2}{c}{Peak (GB)} & \multicolumn{6}{c}{ms per step} \\
    \cmidrule(lr){2-3} \cmidrule(l){4-9}
    & CX & MM & \multicolumn{3}{c}{CharXiv} & \multicolumn{3}{c}{MMMU-Pro} \\
    Method & 64 & 64 & 1 & 16 & 64 & 1 & 16 & 64 \\
    \midrule
    R-KV, $B = 384$ & 21.1 & 21.1 & 41.0 & 38.3 & 41.8 & 42.9 & 38.0 & 41.6 \\
    R-KV, $B = 512$ & 22.8 & 22.8 & 43.1 & 38.8 & 43.4 & 36.2 & 38.1 & 43.4 \\
    \midrule
    R-KV + \method & 20.8 & 21.4 & 48.3 & 60.4 & 103.2 & 48.1 & 68.6 & 152.2 \\
    \quad CPU at step start & 20.8 & 21.4 & 50.8 & 73.1 & 126.8 & 54.7 & 78.8 & 174.4 \\
    \bottomrule
  \end{tabular}
\caption{
  \textbf{Decoding memory and latency on an 80\,GB A100.}
  Qwen3-VL-8B-Thinking; \method uses $B=338$ on both benchmarks.
  Decoding peak memory includes model weights;
  batch sizes are shown in the header.
  CX/MM: CharXiv/MMMU-Pro.
  CPU at step start: sketch scoring without background overlap.
}
\label{tab:cost}
\end{table}

We measure \method's physical memory use and decoding latency with R-KV as the host on one 80\,GB A100.
\cref{tab:cost} shows similar decoding peak memory: at batch 64, \method peaks at 20.8\,GB on CharXiv and 21.4\,GB on MMMU-Pro, against 21.1\,GB for R-KV at $B=384$ and 22.8\,GB at $B=512$.
On-demand reading, however, adds per-step cost that grows with the batch: on CharXiv, \method takes 48.3\,ms against 41.0\,ms for R-KV at $B=384$ at batch 1, and 103.2 against 41.8\,ms at batch 64.

Running sketch selection in the background reduces the time the GPU waits for candidates: at batch 64, it lowers step time from 126.8 to 103.2\,ms on CharXiv and from 174.4 to 152.2\,ms on MMMU-Pro, compared with scoring at step start.

\section{Limitations}
\label{sec:limits}

\method trades decoding time for GPU memory and does not reduce total memory: all visual KV stays in CPU memory, the prefill peak is unchanged, and each decoding step is slower.
It also protects only the image: the host still deletes text for good, so with a weak host such as R-KV, \method stays far below the full cache.
Protecting the image pays off only when a question needs regions that the earlier reasoning did not use.
When the host already keeps what the question needs, as on single-turn MathVerse~\cite{mathverse} questions and short-answer VisDial~\cite{visdial} dialogues, \method with 16-bit keys only matches the host. On small details of HR-Bench 4K~\cite{hrbench} images, every compressed cache, ours included, stays 3.5 to 6.0 points below the full cache.
How much to protect also depends on the conversation: when the reasoning question comes last, a 128-entry reserve does not help on it, while a 256-entry reserve matches the full cache.
Finally, our implementation shelves only images given at prefill, so on MMDU all images enter in the first turn. Our CharXiv conversations are built from an existing benchmark, not collected from users.

\section{Conclusion}
\label{sec:conclusion}

We studied what KV retention does to the image when a reasoning VLM continues a conversation.
The image and the reasoning share one budget, so the growing reasoning crowds the visual entries out of the cache before a later question needs them.
A small visual reserve that the policy cannot evict recovers most of this loss, more than the same slots in a shared budget.
We then proposed \method, a training-free method that runs on top of existing retention policies.
It keeps every visual entry in CPU memory, so the image is never evicted, yet it holds attention slots only while it is read.
In conversations, \method improves DBTrimKV by up to 20.3 points at matched GPU memory, and it still leads by 8.7 points when DBTrimKV uses twice the memory.
It also consistently outperforms a fixed visual reserve, with two retention policies and two models.
On-demand reading adds per-step cost, so \method trades decoding time for a smaller GPU cache.
These results show that the image should be protected from the growing reasoning but read only on demand.

\section*{LLM Usage Statement}
We used LLMs as coding and writing assistants for the experiment and analysis code and the text, and as judges for grading (\texttt{deepseek-v4.1-flash} to check the CharXiv grading, \texttt{gpt-6-luna} and \texttt{gpt-6.1-sol} for MMDU).
The authors decided the research question, the method, the protocol and the conclusions, checked every number against the result files, and take full responsibility for the content.

{
    \small
    \bibliographystyle{ieeenat_fullname}
    \bibliography{refs}
}

\clearpage
\twocolumn[{%
  \centering
  \Large\textbf{Supplementary Material}\par
  \vspace{1.0em}
}]
\appendix
\setcounter{table}{0}
\setcounter{figure}{0}
\renewcommand{\thetable}{S\arabic{table}}
\renewcommand{\thefigure}{S\arabic{figure}}
\raggedbottom

\section{Memory accounting}
\label{sec:s-memory}

\Cref*{eq:memory} of the paper counts the GPU entries per head of \method, a key or a value being half an entry.
With $R_{\max} = B/4$ and $c = 4 R_{\max}$ candidates (rounded down to a multiple of four), \method holds the $B$ entries of the desk and the 4-bit keys of its $c$ candidates, which take $c/8$ entries.
The keys are quantized per channel with a 16-bit minimum and maximum for every group of 128 consecutive visual tokens.
Since candidates can come from any group, the ranges of all $\lceil N_v/128 \rceil$ groups stay on the GPU, one entry per group.
\method therefore holds $B + c/8 + \lceil N_v/128 \rceil$ entries per head for $N_v$ visual tokens.
Averaged over questions, this is 131.6, 258.6 and 385.6 entries for $B = 112$, 225 and 338 on CharXiv and the conversations (671 visual tokens on average), 403.2 at $B = 338$ on MMMU-Pro (2,903) and 430.1 on MMDU (6,358).
For the comparisons at matched GPU memory, we run \method at $B = 320$ on MMMU-Pro and $B = 296$ on MMDU, which hold 383.2 and 383.1 entries.
With full-precision candidate keys, as in the 16-bit row of \cref{tab:s-ondemand}, \method holds $B + c/2$ entries, 384 for $B = 256$.

The shelf, the sketch and the values of the candidates and of parked entries stay in CPU memory.
The sketch takes 54.5 entries per head for an average CharXiv image (671 visual tokens) and 124.2 for an average MMMU-Pro image (2,903).
In the implementation of \cref*{sec:cost}, the CharXiv question of the batch-1 runs (718 prompt tokens) pins 0.14 GB of CPU memory and the MMMU-Pro question (1,512 prompt tokens) 0.27 GB, of which two thirds to three quarters are the keys and values of the visual entries.

Every four steps, a background thread scores the sketch with the query of each layer as soon as the previous step has computed it.
It takes the top $c$ candidates per head and copies their 4-bit codes to the GPU, and each layer waits only for its own candidates.
At every step, the gather kernels of \cref*{sec:memory}, adapted from the open-source kernels of ShadowKV to rows of a single token, read the values and full-precision keys of the borrowed entries directly from pinned CPU memory, so the CPU neither launches nor synchronizes any transfer.
The keys go into a GPU buffer that all layers share and take no persistent GPU memory.

\section{Matched-GPU comparison per benchmark}
\label{sec:s-ondemand}

\Cref{tab:s-ondemand} extends the comparison of \cref*{sec:ondemand} to every benchmark; the paper plots the CharXiv columns.

\begin{table}[h]
  \centering
  \footnotesize
  \setlength{\tabcolsep}{2.4pt}
  \begin{tabular}{@{}lcccccc@{}}
    \toprule
    & \multicolumn{3}{c}{CharXiv} & Conv. & MMMU & MMDU \\
    \cmidrule(lr){2-4}
    GPU entries per head & 128 & 256 & 384 & 384 & 384 & 384 \\
    \midrule
    Best fixed split & 54.0 & 76.7 & 81.0 & 81.9 & 55.3 & 4.6 \\
    \method, prev.\ query & 63.0 & 79.7 & 83.7 & 83.3 & 53.7 & 4.7 \\
    \method, 16-bit keys & 45.3 & 75.7 & 81.3 & 83.9 & 52.7 & 4.8 \\
    \method, 4-bit keys & \textbf{66.7} & \textbf{81.7} & \textbf{84.0} & \textbf{85.0} & \textbf{56.7} & \textbf{4.9} \\
    \bottomrule
  \end{tabular}
  \caption{\textbf{Matched-GPU comparison on all benchmarks} (DBTrimKV as the host; the CharXiv columns are \cref*{fig:ondemand} of the paper).
  Best fixed split: the best split between text budget and reserve that we ran within each budget (128: 96 + 32 and 64 + 64; 256: 192 + 64, 224 + 32 and 128 + 128; 384: 256 + 128, 320 + 64, 352 + 32 and, with 320 entries, 256 + 64).
  The \method rows differ in how the swap scores the candidates: with the previous query and every key in CPU memory, or with the current query and the candidate keys on the GPU in 16 or 4 bits.
  Conv.: turns 2 to 4 of the CharXiv conversations; MMMU: MMMU-Pro; MMDU: judge score.
  With 4-bit keys, \method holds 131.6, 258.6 and 385.6 entries per head on CharXiv and 385.6 on the conversations, and runs at $B = 320$ on MMMU-Pro and $B = 296$ on MMDU to hold 383.2 and 383.1 (\cref{sec:s-memory}).}
  \label{tab:s-ondemand}
\end{table}

\section{Which visual entries \method reads}
\label{sec:s-overlap}

We recorded, at the start of each CharXiv follow-up, which visual entries DBTrimKV still keeps in every layer and head, and which entries \method borrows while it answers.
This run uses full-precision candidate keys at $B = 256$ on the first 64 questions of the evaluation set, of which 56 were recorded in full.
\Cref{tab:s-overlap} gives the results with 95\% confidence intervals from a bootstrap over questions.

\begin{table}[h]
  \centering
  \small
  \setlength{\tabcolsep}{3pt}
  \begin{tabular}{@{}>{\raggedright\arraybackslash}p{3.4cm}cc@{}}
    \toprule
    & All 56 & 11 fixed \\
    \midrule
    Visual entries deleted by the host & 97.9 [97.3, 98.4] & 98.7 \\
    Borrow steps on deleted entries & 91.4 [90.5, 92.3] & 93.5 \\
    \bottomrule
  \end{tabular}
  \caption{\textbf{Overlap between the host and \method} (\%, CharXiv follow-ups, $B = 256$).
  The first row is the 97.9\% of \cref*{sec:crowd} of the paper.
  ``11 fixed'': the questions that \method answers correctly and the host alone does not.}
  \label{tab:s-overlap}
\end{table}

Nine in ten borrow steps fall on entries that the host has deleted.
The host keeps only about 2\% of the visual entries, so most of what \method reads is information that the host would no longer have.
On these 56 questions the host alone answers 39 and \method 49.

\section{Sketch design}
\label{sec:s-sketch}

\Cref{tab:sketch} lists the sketch configurations we compared on the development splits, with DBTrimKV at $B = 256$ as the host; $r$ is the sketch dimension, $b$ its bits per dimension and $c$ the number of candidates.
At equal bytes, 32 dimensions at 4 bits score higher than 16 dimensions at 8 bits on MMMU-Pro, whose larger images also need 256 candidates.

\begin{table}[h]
  \centering
  \small
  \setlength{\tabcolsep}{2.5pt}
  \begin{tabular}{@{}cccccc@{}}
    \toprule
    $r$ & $b$ & $c$ & refresh & CharXiv & MMMU-Pro \\
    \midrule
    32 & 8 & 64 & every step & 71.9 & 54.7 \\
    32 & 8 & 256 & every 4 steps & 73.4 & 54.7 \\
    16 & 8 & 256 & every 4 steps & 78.1 & 48.4 \\
    32 & 4 & 128 & every 4 steps & 82.8 & 45.3 \\
    32 & 4 & 256 & every 4 steps & 75.0 & 54.7 \\
    \midrule
    \multicolumn{4}{@{}l}{DBTrimKV, $B = 256$ / 512} & 67.2 / 73.4 & 45.3 / 54.7 \\
    \bottomrule
  \end{tabular}
  \caption{\textbf{Sketch design on the development splits} (64 questions each, $B = 256$).
  The fifth configuration is the one we use.}
  \label{tab:sketch}
\end{table}

We chose how to store the candidate keys (\cref*{tab:ablation}b) on the development splits by the highest mean accuracy over the six cells of \cref{tab:s-variants}, which selects 4-bit keys.

\begin{table}[h]
  \centering
  \small
  \setlength{\tabcolsep}{3pt}
  \begin{tabular}{@{}lccccccc@{}}
    \toprule
    & \multicolumn{3}{c}{CharXiv} & \multicolumn{3}{c}{MMMU-Pro} & \\
    \cmidrule(lr){2-4}\cmidrule(lr){5-7}
    Candidate keys & 128 & 256 & 384 & 128 & 256 & 384 & Mean \\
    \midrule
    Previous query (CPU) & 65.6 & 76.6 & 79.7 & 31.2 & 45.3 & 54.7 & 58.9 \\
    16-bit keys (GPU) & 46.9 & 71.9 & 75.0 & 35.9 & 50.0 & 54.7 & 55.7 \\
    4-bit keys (GPU) & 65.6 & 78.1 & 73.4 & 43.8 & 51.6 & 56.2 & \textbf{61.5} \\
    \bottomrule
  \end{tabular}
  \caption{\textbf{Storage of the candidate keys on the development splits} (64 questions each, DBTrimKV as the host).
  Columns: GPU entries per head, matched across rows; on MMMU-Pro, the 4-bit rows hold about 20 more for the minima and maxima of the visual groups.}
  \label{tab:s-variants}
\end{table}

\section{Agreement between the implementations}
\label{sec:s-agree}

The accuracies in the paper come from our evaluation framework, which applies the selection of \method as masks over a full cache.
The cost measurements come from a physical implementation on R-KV that stores only what \method keeps (\cref*{sec:cost}).
We compare the two on eight CharXiv reasoning questions and four MMMU-Pro questions, with R-KV at $B = 338$ as the host, 4-bit candidate keys and greedy decoding.

\paragraph{Free generation.}
Each implementation generates up to 512 tokens on its own, and we record the first token at which the output of the physical implementation differs from that of the evaluation framework.
The median first difference is at token 179 on CharXiv and 90.5 on MMMU-Pro (\cref{tab:s-free}).
For comparison, two implementations of R-KV alone at $B = 256$, the mask-based one in our framework and a physical one, have their median first difference at token 165.5 on CharXiv and 171.5 on MMMU-Pro.
The tokens are identical on all twelve questions whether the CPU work runs in the background or at the start of each step, and whether or not each layer is captured as a CUDA graph.

\begin{table*}[t]
  \centering
  \small
  \setlength{\tabcolsep}{3.5pt}
  \begin{tabular}{@{}lrrrrrrrrr@{\hspace{12pt}}rrrrr@{}}
    \toprule
    & \multicolumn{9}{c}{CharXiv question} & \multicolumn{5}{c}{MMMU-Pro question} \\
    \cmidrule(r{12pt}){2-10}\cmidrule{11-15}
    Method & 9 & 10 & 16 & 20 & 21 & 24 & 25 & 27 & Median & Pha.\ 287 & His.\ 76 & Psy.\ 35 & Lit.\ 19 & Median \\
    \midrule
    \method, $B = 338$ & 375 & 361 & 362 & 154 & 165 & 41 & 193 & 55 & 179 & 34 & 121 & 82 & 99 & 90.5 \\
    R-KV alone, $B = 256$ & 252 & 147 & 307 & 141 & 253 & 43 & 184 & 103 & 165.5 & 148 & 195 & 255 & 80 & 171.5 \\
    \bottomrule
  \end{tabular}
  \caption{\textbf{First token at which free generation differs}, per question, greedy decoding up to 512 tokens, R-KV as the host.
  The first row compares the physical implementation of \method with the evaluation framework; the second compares the physical and the mask-based implementations of R-KV alone.
  Questions are numbered by their CharXiv index and by their MMMU-Pro subject and index (Pha.: Pharmacy, His.: History, Psy.: Psychology, Lit.: Literature).}
  \label{tab:s-free}
\end{table*}

\paragraph{Forced generation.}
To separate rounding effects from real differences, we feed the output of the evaluation framework token by token to the physical implementation and record every step at which it would choose a different next token, together with the gap between its own top two logits.
On 3,943 forced CharXiv steps and 2,044 MMMU-Pro steps, it chooses differently at 41 and 31 steps.
At 48 of these 72 steps its top two logits are tied in bf16, and at all 72 they are at most 0.625 apart.

\paragraph{Hardware.}
The accuracy runs were spread over the nodes of a shared cluster, which include nodes with 80 GB A800 GPUs and nodes with 80 GB A100 GPUs.
Rerunning 52 CharXiv conversations with DBTrimKV alone and 48 CharXiv follow-ups with R-KV alone (both at $B = 384$) on an A800 node reproduces every generated token of the original runs, which used A100 nodes (28 and 24) and A800 nodes (24 and 24).

\section{Evaluation protocol}
\label{sec:s-protocol}

\paragraph{Decoding.}
Unless stated otherwise, every method runs Qwen3-VL-8B-Thinking with its recommended sampling (temperature 1.0, top-p 0.95, top-k 20) and a fixed per-question seed, so that two methods compared on the same question differ only in their cache.
Generation stops at 16,384 tokens on the single-turn benchmarks and in the full-cache first turn of the CharXiv follow-ups, at 20,480 tokens on the follow-ups including the fed-in first turn, at 12,288 tokens per turn and 32,768 in total in the CharXiv conversations, and at 4,096 tokens per turn on MMDU; an answer that hits a limit counts as wrong.
All methods, including DBTrimKV with its released gates, run in the same harness and software environment.

\paragraph{Benchmarks.}
The CharXiv follow-ups use 300 charts from the validation split of CharXiv, stratified by subject; the first turn is the reasoning question for the chart, and the second turn asks one of the official descriptive questions about the same chart.
The CharXiv conversations chain the descriptive questions of each chart after its reasoning question into one five-turn conversation in which every method generates every turn itself; we score turns 2 to 4, since the fifth question cannot be answered from the chart (\cref{sec:s-fiveturn}).
We use 300 MMMU-Pro Vision questions, stratified by subject.
MMDU has 110 public multi-image conversations (1,645 questions, median 16 turns); every method generates every turn, with all images in the first turn.
With Qwen3-VL-4B-Thinking, we leave out the 5 follow-ups whose first answer from the full cache hits the limit, which leaves 295.

\paragraph{Baselines and configuration.}
The hosts run alone at budgets from 128 to 1024 entries per head, and DSTP with R-KV keeps a fixed visual share of the budget (75\%, chosen on a development split of MathVerse).
GPU entries per head are averaged over questions.
The configuration of \method, including the 4-bit storage of the candidate keys (\cref{sec:s-sketch}), was chosen on development splits of 64 questions per benchmark, disjoint from the evaluation sets.

\begin{table}[t]
  \centering
  \small
  \setlength{\tabcolsep}{3.5pt}
  \begin{tabular}{@{}lcccccc@{}}
    \toprule
    Method & 1 & 2 & 3 & 4 & 5 & 2--4 \\
    \midrule
    Full cache & 50.7 & 85.0 & 88.0 & 89.0 & 79.7 & 87.3 \\
    DBTrimKV, $B = 256$ & 45.0 & 66.7 & 59.0 & 62.3 & 45.3 & 62.7 \\
    DBTrimKV, $B = 384$ & 46.3 & 71.3 & 69.7 & 68.0 & 51.3 & 69.7 \\
    DBTrimKV, $B = 512$ & 43.3 & 71.7 & 72.7 & 72.0 & 48.3 & 72.1 \\
    \midrule
    + reserve ($V = 128$) & 46.7 & 82.0 & 82.0 & 81.7 & 60.0 & 81.9 \\
    + \method, 4-bit keys & 46.3 & 83.3 & 84.7 & 87.0 & 68.0 & 85.0 \\
    \bottomrule
  \end{tabular}
  \caption{\textbf{Accuracy (\%) by turn} in the 300 five-turn CharXiv conversations (Qwen3-VL-8B-Thinking, DBTrimKV as the host; the reserve holds 384 and \method 386 GPU entries per head).
  Rows with + add the reserve or \method to DBTrimKV.
  Of each method's 1,200 answers in turns 2 to 5, one each for DBTrimKV at $B = 512$ and \method reaches the 12,288-token limit of a turn, and none for the other methods.}
  \label{tab:s-fiveturn}
\end{table}

\section{Grading}
\label{sec:s-grading}

\paragraph{CharXiv.}
We score CharXiv with a rule that approximates the official grader.
To check it, an LLM judge (\texttt{deepseek-v4.1-flash}) re-scored the follow-up answers of the four configurations in \cref{tab:s-cxjudge} with the official descriptive grading prompts and in-context examples, five answers per request as in the official code.
\Cref{tab:s-cxjudge} compares the two.

\begin{table}[h]
  \centering
  \small
  \setlength{\tabcolsep}{4pt}
  \begin{tabular}{@{}lccc@{}}
    \toprule
    Method & Rule & LLM judge & Agreement \\
    \midrule
    Full cache & 85.7 & 85.7 & 98.7 \\
    DBTrimKV, $B = 256$ & 61.3 & 61.3 & 98.0 \\
    DBTrimKV, $B = 512$ & 73.0 & 73.3 & 99.0 \\
    \method, 16-bit keys, $B = 256$ & 81.3 & 82.3 & 97.7 \\
    \bottomrule
  \end{tabular}
  \caption{\textbf{CharXiv follow-ups scored by our rule and by an LLM judge with the official prompts} (\%).
  Agreement: answers that both score the same.}
  \label{tab:s-cxjudge}
\end{table}

\paragraph{MMDU.}
We copy the scoring prompt of the official MMDU evaluation verbatim and score each turn as text (question, reference answer and model answer) with the judge's default sampling settings; an answer that hits the 4,096-token turn limit receives a score of 1.
We chose the judge on 56 answers from two conversations, where \texttt{gpt-6-luna} agreed best with a stronger reference judge (\texttt{gpt-6.1-sol}) among three candidates (Pearson 0.81).
On 487 answers from nine conversations, the two judges have Pearson and Spearman correlations of 0.88 and a mean absolute difference of 0.51, order 84\% of answer pairs the same way, and rank the methods the same way.
The released DBTrimKV gates list MMDU among their training sources; their public training recipe downloads only the MMDU-45K training conversations, not the benchmark conversations we evaluate on.
The other listed sources, R1-Onevision, M4-Instruct, LLaVA-Video-178K and OpenR1-Math-220k, are not among our benchmarks.

\section{Five-turn conversations by turn}
\label{sec:s-fiveturn}

\Cref{tab:s-fiveturn} gives the accuracy of each turn of the CharXiv conversations.
Turn 1 is the reasoning question, which every method answers itself, so each method's follow-ups build on its own first answer.
Turns 2 to 5 are the four descriptive questions of the chart in dataset order.
The reference answer of the fifth is ``Not Applicable'' in all 300 conversations, so turn 5 measures whether the model declines to answer rather than how deep the conversation is; the paper therefore reports turns 2 to 4.
On these turns, \method with 4-bit keys leads DBTrimKV at $B = 512$ by 11.6 to 15.0 points and the reserve by 1.3 to 5.3 points.

\section{Code}
We will release the code, including \method on both hosts, the baselines, the reserve and its controls, the construction of the CharXiv follow-ups and conversations, the grading and statistics, and the physical implementation with its kernels and tests.

\end{document}